\documentclass[10pt, conference]{IEEEtran}
\IEEEoverridecommandlockouts

\usepackage{adjustbox}
\usepackage{subcaption}
\usepackage{caption}
\usepackage{graphicx} 
\usepackage{stfloats}
\usepackage{booktabs}
\usepackage{multirow}
\usepackage{optidef}
\usepackage{makecell}

\usepackage[percent]{overpic}
\usepackage{graphicx}
\usepackage{amsmath,amssymb,amsfonts}
\usepackage{float}
\usepackage{multicol}
\usepackage{lipsum} 
\usepackage{textcomp}
\usepackage{xcolor}
\usepackage{breakurl}
\usepackage{xurl}
\usepackage{longtable} 
\usepackage{array}     
\usepackage{verbatim}
\usepackage[ruled, vlined, linesnumbered, boxed]{algorithm2e}
\usepackage{algpseudocode}
\usepackage{cite}

\title{Deep Reinforcement Learning-driven Edge Offloading for Latency-constrained XR pipelines\thanks{This material is based upon work supported by the National Science Foundation (NSF) under Award Number CNS-1943338.}}

\author{
Sourya Saha, Saptarshi Debroy\\
City University of New York\\ 
Emails: \textit{ssaha2@gradcenter.cuny.edu, saptarshi.debroy@hunter.cuny.edu}

}

\begin{document}

\maketitle
\thispagestyle{empty}
\pagestyle{empty}
\maketitle

\begin{abstract}
Immersive extended reality (XR) applications introduce latency-critical workloads that must satisfy stringent real-time responsiveness while operating on energy- and battery-constrained devices, making execution placement between end devices and nearby edge servers a fundamental systems challenge. Existing approaches to adaptive execution and computation offloading typically optimize average performance metrics and do not fully capture the sustained interaction between real-time latency requirements and device battery lifetime in closed-loop XR workloads. In this paper, we present a battery-aware execution management framework for edge-assisted XR systems that jointly considers execution placement, workload quality, latency requirements, and battery dynamics. We design an online decision mechanism based on a lightweight deep reinforcement learning policy that continuously adapts execution decisions under dynamic network conditions while maintaining high motion-to-photon latency compliance. Experimental results show that the proposed approach extends the projected device battery lifetime by up to 163\% compared to latency-optimal local execution while maintaining over 90\% motion-to-photon latency compliance under stable network conditions. Such compliance does not fall below 80\% even under significantly limited network bandwidth availability, thereby demonstrating the effectiveness of explicitly managing latency–energy trade-offs in immersive XR systems.
\end{abstract}

\begin{IEEEkeywords}
Extended reality, edge offloading, deep reinforcement learning, latency optimization, energy optimization.
\end{IEEEkeywords}

\section{Introduction}
\label{sec:intro}

Recent advances in immersive extended reality (XR) applications, including virtual reality (VR), augmented reality (AR), and mixed reality (MR), are driving a new class of interactive and latency-sensitive workloads in distributed computing systems. These applications rely on continuous processing of sensed data and rendering pipelines to maintain real-time responsiveness, imposing stringent latency requirements typically of the order of tens of milliseconds. At the same time, XR devices operate under tight energy and battery limitations, with restricted on-device compute and thermal budgets. To meet growing compute demands without sacrificing device lifetime, execution is increasingly shifted toward nearby {\em edge servers}, typically placed close to end-users~\cite{effect,effect-dnn}.
This architectural shift exposes a fundamental tension between responsiveness and energy efficiency in latency-critical systems~\cite{survey}. 

\begin{figure}[t]
    \centering
    \includegraphics[width=\linewidth]{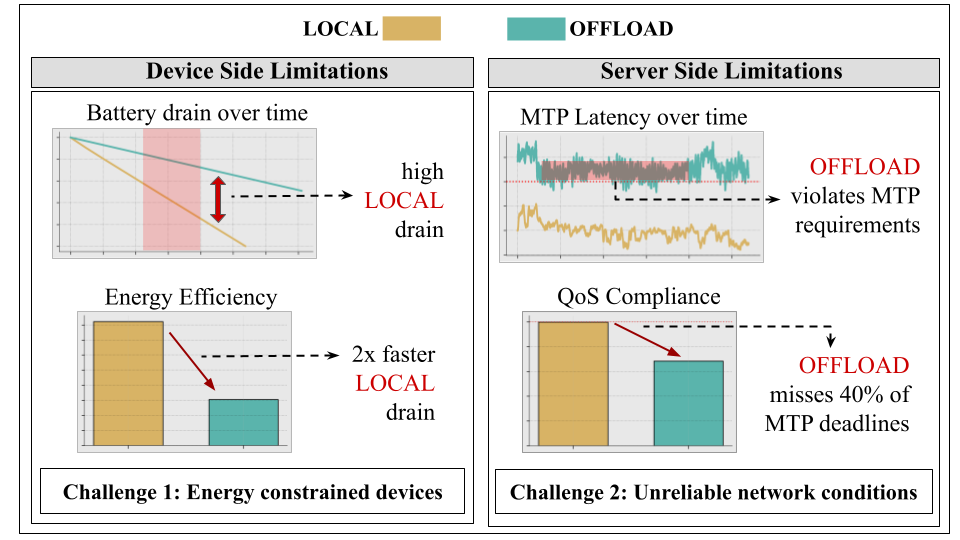}
    \caption{The non-trivial conflict between Local vs Remote Offloading of XR workloads. 
    }
    \label{fig:fig1}
    \vspace{-0.2in}
\end{figure}

Addressing this tension, i.e., satisfying real-time responsiveness while prolonging end-device lifetime, 
is critical to enable sustained and responsive XR experiences on resource-constrained devices.
This is because latency, energy consumption, and execution placement in XR applications are tightly intertwined. In XR, motion-to-photon (MTP) latency, i.e., \emph{the delay for a user movement being reflected accurately in the virtual scene}, must typically be less than 20 ms (approximately) to preserve perceptual stability and user comfort \cite{bassbouss2016high,asim2025generationimmersiveapplications5g}, leaving little tolerance for additional delay. Performing computation locally avoids communication latency but imposes sustained high computational load on the device, rapidly draining battery. Conversely, shifting computation away from the device reduces local energy expenditure but introduces additional communication and processing delays that directly contribute to end-to-end MTP latency. These effects interact in non-trivial ways: decisions about execution placement simultaneously influence power consumption, latency, and sensitivity to network conditions. We illustrate this trade-off in Fig. \ref{fig:fig1}, which shows that purely local execution leads to rapid battery depletion, while offloaded execution can cause MTP latency to exceed acceptable bounds, highlighting why neither extreme is sufficient in isolation.

\emph{Immersive XR workloads differ fundamentally from other edge-native video processing applications as the former form closed-loop, stateful pipelines.} In such systems, performance is governed by short-term, continuous behavior rather than long-term average metrics, as delayed or inconsistent pose updates cannot be masked through buffering and may immediately degrade user perception. Against this backdrop, prior work has extensively examined task offloading and energy--latency trade-offs for mobile applications through analytical and optimization-based formulations \cite{ZhangDebroy2020, ZhangEtAl2021,xiaojie-short}, as well as heuristic and learning-driven decision policies designed to adapt execution placement under dynamic network conditions \cite{TangWong2020}.

Computation offloading for immersive XR workloads presents challenges that are not fully addressed by existing offloading formulations. Many approaches optimize average latency or throughput and rely on simplified energy abstractions, without prioritizing real-time latency compliance or explicitly modeling battery lifetime as an important system objective \cite{Dong2024TaskOffloadingSurvey, Alwabel2025LatencyAware}. Empirical XR studies show that although offloading can reduce local power consumption, responsiveness remains highly sensitive to network latency and short-term variability, with execution latency often treated as an optimization objective rather than a deterministic guarantee \cite{Gao2025XRgo, Ometov2023EdgeEnergy}. Recent adaptive and partial offloading methods introduce dynamic resource allocation, yet primarily focus on short-term efficiency metrics rather than sustained operation under coupled latency and energy constraints \cite{Duru2025ResourceAllocationXR,yu-drl-xr-mobile-edge}. As a result, existing formulations fail to capture the interaction between execution placement, workload quality, real-time constraints, and battery dynamics that governs closed-loop XR execution.

In this paper, we present a novel framework for battery-aware and real-time execution management in edge-assisted XR applications. We formalize the interaction between computation placement, workload quality, latency requirements, and battery dynamics into a unified model that captures the challenges faced by resource-constrained XR devices operating within closed-loop perception pipelines and stringent MTP latency bounds. Building on this formulation, we design an online decision mechanism that continuously selects execution configurations to extend device lifetime while ensuring real-time responsiveness under fluctuating network conditions. Unlike prior approaches that optimize average performance or rely on static heuristics, our framework prioritizes MTP latency while jointly optimizing battery lifetime as a system objective. Instead of assuming deterministic network conditions, our approach learns policies that aims to maintain high MTP compliance under stochastic network dynamics. In particular, we instantiate this design using a lightweight deep reinforcement learning (DRL)–based control policy and integrate it into a full end-to-end prototype, enabling practical deployment and evaluation under realistic workloads and network dynamics. While motivated by XR, the proposed framework generalizes to other latency-critical, closed-loop mobile applications that rely on proximity-based computation to balance responsiveness and energy efficiency.

We evaluate the proposed framework using an end-to-end implementation built on ILLIXR \cite{illixr}, an open immersive computing research platform, which enables controlled experimentation across execution placement, network conditions, and workload configurations. Our evaluation focuses on system-level metrics central to immersive performance and device sustainability, including MTP latency, instantaneous power draw, and resulting battery lifetime. By comparing adaptive execution decisions against fixed and non-adaptive configurations under representative network conditions, we demonstrate how coordinated execution and quality control enable sustained real-time responsiveness while reducing energy consumption. 
The results show that the proposed framework achieves a favorable trade-off between battery lifetime and latency compliance, extending projected battery life by approximately 163\% compared to latency-optimal local execution while maintaining MTP compliance above 90\% under stable network conditions, and over 80\% MTP compliance at even the least of evaluated bandwidths. This highlights the effectiveness of explicitly modeling and managing latency–energy trade-offs in latency-constrained systems.

The remainder of the paper is organized as follows. Section \ref{sec:related_works} reviews related work. 
Section \ref{sec:system_model} introduces the system model and formalizes 
our framework. Section \ref{sec:rl_model} presents the DRL–based decision model. 
Section \ref{sec:eval} describes the experimental evaluation and evaluation results.
Section \ref{sec:conclusion} concludes the paper and outlines directions for future work.

\section{Related Work}
\label{sec:related_works}

High energy consumption is a primary concern for mobile and embedded devices, where limited battery capacity and thermal budgets restrict sustained on-device computation. As modern applications increasingly incorporate compute-intensive processing, executing all workloads locally can rapidly drain battery resources. This has motivated the use of edge offloading, which shifts computation toward nearby infrastructure with greater compute and power availability.
Early energy-aware task offloading formulations further demonstrate that transmission energy and delay can offset compute savings, motivating joint consideration of computation placement and communication cost \cite{Dong2024TaskOffloadingSurvey,ZhangDebroy2020}.

Beyond full task migration, selective and partial offloading strategies have been explored to reduce communication overhead and improve average energy efficiency under latency constraints. These approaches include optimization-based fine-grained task partitioning and lightweight local preprocessing \cite{ZhangDebroy2020, ZhangEtAl2021, effect, effect-dnn}, to learning-based formulations to adapt offloading decisions \cite{TangWong2020,Ometov2023EdgeEnergy,infer-edge, motaharevec}.
However, much of the existing offloading literature remains focused on per-task or average-case optimization and tolerates occasional delay violations. Such assumptions become unrealistic for tightly coupled immersive pipelines (e.g., XR), where responsiveness must be preserved with high temporal consistency rather than optimized only for long-term expectation. This motivates closer examination of edge-assisted execution under stringent real-time requirements.

Unlike many mobile applications that can tolerate variable execution delay, immersive XR workloads impose continuous and stringent MTP latency requirements at the millisecond scale that fundamentally alter the offloading design space \cite{Dixit2023MTP}. 
Recent systems research illustrates both the promise and fragility of edge-assisted execution for XR. Several works show that offloading computationally intensive tracking or rendering components can substantially reduce headset-side power consumption, validating offloading as an effective mechanism for improving device energy efficiency \cite{RemoteVIO2025, Zhu2022EdgeVR, Gao2025XRgo, xiaojiesec2023}. 
At the same time, these work demonstrate that XR performance remains highly sensitive to end-to-end latency and temporal variability. 
Detailed measurements reveal that short-term effects such as packet batching, congestion intervals, or scheduling artifacts can delay complete frame delivery and push MTP latency beyond acceptable bounds, even when average network delay remains low \cite{Casasnovas2024EdgeCloudVR,Yeregui2024EdgeXR, Okafor2024MultiUserXR,BELLALTA2026104391}.

Collectively, these findings underscore a fundamental distinction between XR workloads and generic edge-assisted applications. XR performance is governed by short-term and worst-case execution behavior rather than average-case metrics. Consequently, offloading strategies that focus solely on mean latency or short-term energy savings are insufficient for sustained immersive operation. Effective XR execution management must therefore jointly account for stringent real-time latency requirements, energy and battery dynamics, and runtime variability.
Further, XR applications execute continuously and react directly to user motion, causing execution and offloading decisions to influence perceptual output on short timescales. These effects cannot be averaged over long windows, making transient instability immediately visible to the user. Existing learning-based offloading approaches typically optimize expected or short-horizon objectives and treat energy as an instantaneous or per-task cost signal, without explicitly modeling how execution decisions accumulate over sustained device operation. These limitations motivate execution management frameworks that adapt learning-based control to XR’s continuous execution dynamics and evolving energy state. In the next section, we introduce a system model that captures these interactions and forms the foundation for our adaptive execution approach.

\section{System Model}
\label{sec:system_model}

We present a system model for edge-assisted execution in interactive XR systems, capturing the interplay between computation placement, workload configuration, energy consumption, and real-time responsiveness under dynamic conditions.

\begin{figure}[t]
  \centering
  \includegraphics[width=1\columnwidth]{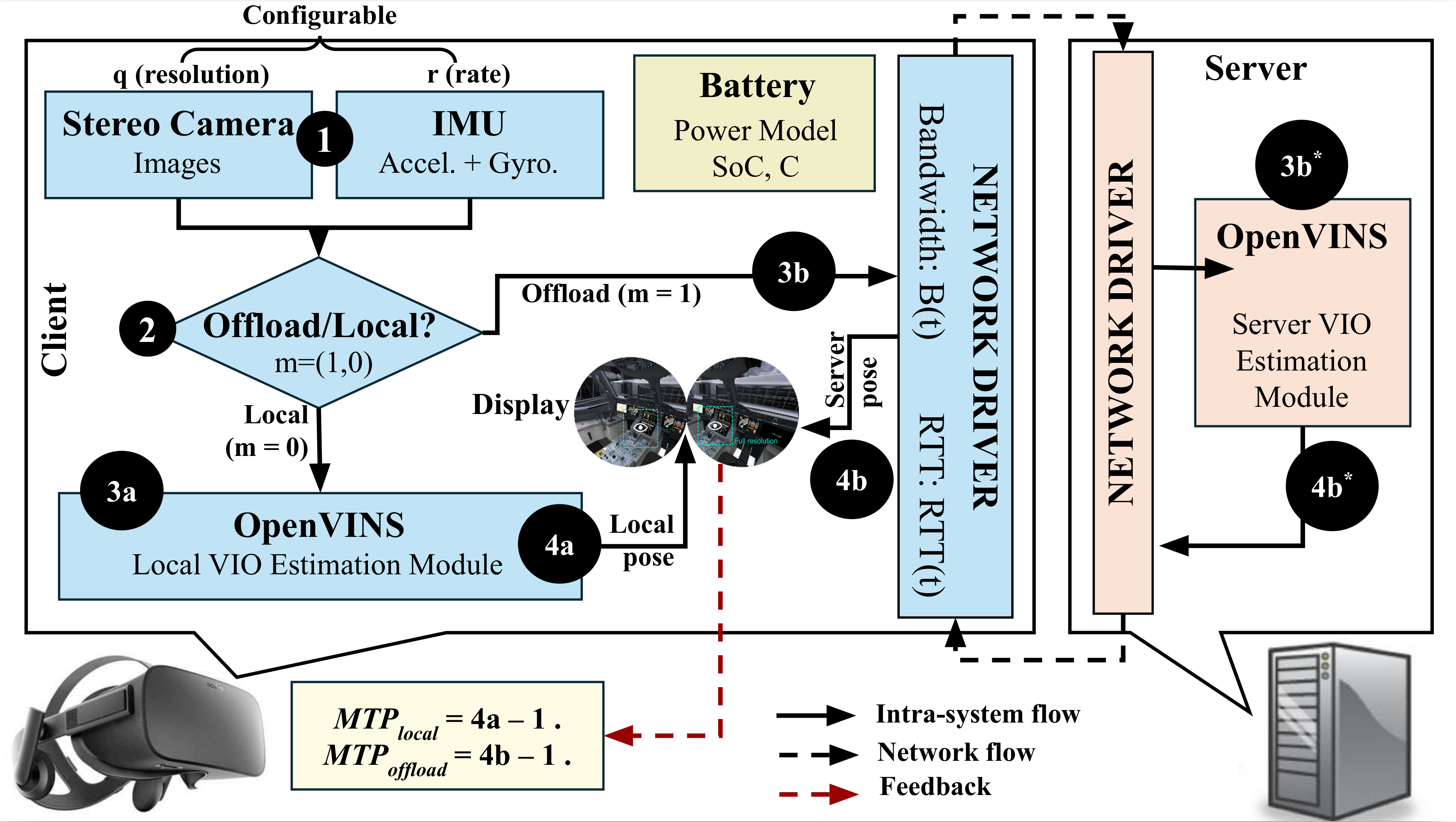}
  \caption{Overview of the edge-assisted interactive XR system model}
  \label{fig:setup}
  \vspace{-0.2in}
\end{figure}

\vspace{-0.1in}
\subsection{System Architecture and Execution Modes}
\label{sec:execution-mode}

We consider an interactive XR system consisting of a resource-constrained client device and a nearby compute server connected via a network. The client operates under limited computational capacity and a finite battery budget, while the server is assumed to have sufficient compute resources and no energy constraints. The client continuously executes a perception-driven processing pipeline to support real-time XR interaction, and selected components of this pipeline may be offloaded to the server to reduce on-device load.

We use visual-inertial odometry (VIO) as a motivating workflow in our system model. VIO continuously fuses visual and inertial data to estimate device motion and operates persistently on the critical path of XR execution. As a result, its execution latency and energy consumption directly influence end-to-end interaction responsiveness.

The system supports two execution modes: i) In local execution mode $(m = 0)$, all processing is performed on the client device; and ii) In offloaded execution mode $(m = 1)$, the client transmits the required sensor data or intermediate representations to the server, which performs the compute-intensive processing and returns the resulting outputs to the client over the network. Fig. \ref{fig:setup} provides an abstract view of these execution modes and their interaction with network delay and energy consumption; the figure illustrates logical execution paths rather than a specific software architecture.

To enable runtime adaptation, the system exposes three configurable parameters that jointly define an execution configuration. First, a visual quality level $q \in \{0,1,2\}$ controls the fidelity of visual data processed and transmitted. Second, an inertial processing rate $r \in \{0,1,2\}$ governs the effective rate at which inertial measurements are incorporated into the processing pipeline. Third, the execution mode $m \in \{0,1\}$ determines whether processing is performed locally or offloaded. The tuple $(q,r,m)$ is selected at runtime and remains fixed over the corresponding decision interval.

This abstraction captures a range of execution strategies spanning fully local processing, full offloading, and intermediate configurations that trade off computation, communication, and energy usage.

\subsection{Performance Model: VR Latency}

The primary performance metric in immersive XR systems is MTP latency, defined as the elapsed time between a user head movement and the corresponding visual update presented to the display. MTP latency directly impacts perceptual stability and user comfort, since excessive and constantly delayed updates can lead to disorientation and motion sickness. Consequently, XR systems must maintain MTP latency within tight application-dependent bounds with high temporal consistency to satisfy minimum acceptable quality of experience. In practice, XR platforms commonly target MTP latencies on the order of a few tens of milliseconds; the specific bound used for evaluation in our work is specified later in Section~\ref{sec:eval}.

We measure MTP latency as the time difference between sensor capture and pose availability,
\begin{equation}
\text{MTP} = t_{\text{pose}} - t_{\text{capture}}
\label{eq:mtp}
\end{equation}
where $t_{\text{capture}}$ denotes the timestamp at which a visual frame is captured on the client device and $t_{\text{pose}}$ denotes the timestamp at which the corresponding pose estimate becomes available for display. This definition captures the end-to-end latency of the XR perception loop and is independent of where computation is performed.

For local execution, MTP latency includes sensor capture, on-device perception processing, and rendering delay. For offloaded execution, MTP latency additionally includes encoding overhead, network transmission from the client to the server, server-side processing, return transmission of the pose estimate, and decoding overhead. As a result, offloading decisions introduce a direct dependence between network conditions and end-to-end latency.

The network exhibits time-varying behavior that significantly influences the feasibility of offloaded execution. We characterize the network at time $t$ using two parameters: the available uplink bandwidth $B(t)$ and the round-trip time $\text{RTT}(t)$. Both parameters vary due to congestion, interference, and user mobility. During offloaded execution, visual and inertial data are transmitted from the client to the server, while the server returns a compact pose estimate to the client. Since the returned pose message is small relative to the uplink payload, its transmission time is neglected.

The network-induced delay associated with offloading is therefore modeled as:
\begin{equation}
t_{\text{net}}(q,t) = \frac{D_{\text{base}} \cdot \phi(q)}{B(t)} + \text{RTT}(t),
\label{eq:network_time}
\end{equation}
where $D_{\text{base}}$ denotes the data size corresponding to the highest visual quality level, and $\phi(q)$ is a quality-dependent scaling factor capturing the relative data volume at quality level $q$. The first term represents uplink serialization delay, while the $\text{RTT}(t)$ captures the round-trip propagation and queuing delay of the request-response interaction between client and server.

Lower quality levels reduce data volume, thereby decreasing both network transmission time and processing cost. This coupling between execution configuration, network conditions, and end-to-end latency motivates adaptive control strategies that jointly account for computation placement and workload quality under dynamic operating conditions.

\subsection{Power Model}

Battery lifetime is a key constraint for mobile XR devices and is directly influenced by execution decisions. We model energy usage using proxy signals that capture how different execution configurations affect computational load. The instantaneous power consumption of the client is modeled as:
\begin{equation}
P_{\text{client}} =
P_{\text{base}} +
w_{\text{proc}}\,P_{\text{proc}}(q, r, m) +
w_{\text{cpu}}\, P_{\text{cpu}} + w_{\text{gpu}}\,
P_{\text{gpu}}
\label{eq:power_total}
\end{equation}
where each term represents a proxy for energy-related system activity and $w_{\cdot}$ denote weighting factors.
The baseline power $P_{\text{base}}$ represents power drawn by always-on system components such as the display, operating system, memory, and idle hardware, and is independent of execution decisions.
The processing-related term $P_{\text{proc}}(q, r, m)$ captures the compute intensity induced by the XR perception workload under execution configuration $(q, r, m)$. This term is explicitly decision-dependent and correlates directly with processing time, which increases with higher image quality and higher inertial sampling rates. We model this contribution as:
\begin{equation}
P_{\text{proc}}(q, r, m) =
\frac{t_{\text{proc}}(q, r, m)}{\tau_{\text{frame}}}
\cdot
\text{TDP}_{\text{proc}},
\label{eq:power_proc}
\end{equation}
where $t_{\text{proc}}(q, r, m)$ denotes the processing time associated with the workload, $\tau_{\text{frame}}$ is the target frame duration corresponding to the inverse of the frame rate (e.g., a 20~Hz image capture pipeline yields $\tau_{\text{frame}} = 50$~ms), and $\text{TDP}_{\text{proc}}$ represents the processor thermal design power.
The terms $P_{\text{cpu}}$ and $P_{\text{gpu}}$ represent auxiliary indicators of system activity that are observable on the evaluation platform but are only weakly influenced by changes in workload quality or execution mode. As a result, variations in these signals do not reliably reflect how energy consumption would scale on a resource-constrained XR headset.
To account for this discrepancy, we apply weighting factors in Eqn. \eqref{eq:power_total}.
For simplicity, power consumption due to image acquisition and inertial sensing is absorbed into the baseline term, as it does not affect the relative trade-offs considered in this work.

\subsection{VR Headset Battery Model}

We model the battery behavior of an VR headset using a simple energy balance formulation that relates power consumption to battery depletion over time. The battery is characterized by a total energy capacity $C$ (Wh) and a state of charge $\text{SoC}(t) \in [0,100]$, expressed as a percentage of remaining capacity.

The state of charge evolves according to
\begin{equation}
\text{SoC}(t + \Delta t) =
\text{SoC}(t) -
\frac{P_{\text{client}}(t)\,\Delta t}{C} \cdot 100,
\label{eq:battery_dynamics}
\end{equation}
where $P_{\text{client}}(t)$ is the instantaneous power consumption defined in Eqn. \eqref{eq:power_total}, and $\Delta t$ denotes the time step over which energy consumption is accumulated. This formulation captures the direct relationship between power draw and battery depletion: higher instantaneous power leads to faster reduction in available battery charge.

Under the simplifying assumption of a constant power draw $P$, the remaining battery lifetime can be expressed as
\begin{equation}
T_{\text{remain}} =
\frac{\text{SoC}(t)}{100}
\cdot
\frac{C}{P},
\label{eq:battery_lifetime}
\end{equation}
where $T_{\text{remain}}$ is measured in hours. This expression provides an intuitive baseline for understanding how power consumption impacts device lifetime.

In practice, however, power consumption varies over time as execution configurations change in response to workload characteristics and system conditions. As a result, battery lifetime is determined by the cumulative effect of time-varying power draw rather than a fixed operating point. This dynamic behavior motivates adaptive execution strategies that seek to extend battery lifetime while satisfying real-time performance constraints.

\subsection{Optimization Problem Formulation}

The fundamental challenge in edge-assisted interactive XR is to dynamically configure the XR perception pipeline to maximize battery lifetime while ensuring that MTP latency remains within acceptable bounds. This optimization must be performed online under time-varying network conditions, uncertain future states, and the interaction of multiple control parameters.

First, we prioritize MTP latency. Rather than assuming deterministic satisfaction of a hard latency bound under stochastic network conditions, we quantify the extent to which latency exceeds the target threshold and penalize such violations during optimization. We define:

\begin{equation}
V(t) = \max\left(0,\; \frac{\text{MTP}_{\text{cam}}(q(t),\, m(t),\, B(t)) -     \tau_{\text{MTP}}}{\tau_{\text{MTP}}}\right)
\label{eq:mtp_violation}
\end{equation}
where $V(t) $ denotes the normalized violation magnitude, expressing fractional MTP latency violation beyond the threshold $\tau_{\text{MTP}}$. Normalizing by the threshold ensures that the penalty term for MTP in the optimization objective remains interpretable independently of the specific threshold value.
Here, MTP latency depends on both controllable parameters, such as workload quality $q(t)$ and execution mode $m(t)$, and uncontrollable network conditions, including time-varying bandwidth $B(t)$ and round-trip time. Degradation in network conditions can cause offloaded execution to violate latency requirements, requiring rapid adaptation through quality reduction or local execution.  
Note that the effect of IMU sampling rate $r(t)$ on MTP latency is captured implicitly through its impact on processing time, rather than as a separate latency component.

We formulate the system objective as maximizing the operational lifetime of the device while penalizing MTP latency violations. Let $T_{\text{battery}}$ denote the stopping time at which the battery state of charge reaches zero. Using the violation magnitude $V(t)$ defined in Eq.~(\ref{eq:mtp_violation}), the optimization problem becomes:

\begin{equation}
\begin{aligned}
\max_{q(t), r(t), m(t)} \quad & T_{\text{battery}} - \lambda \sum_t V(t) \\
\text{subject to} \quad
& \frac{d\text{SoC}(t)}{dt}
= -\frac{P_{\text{client}}(t)}{C} \cdot 100, \\
& \text{SoC}(0) = \text{SoC}_0,\quad \text{SoC}(T_{\text{battery}})=0.
\end{aligned}
\label{eq:optimization}
\end{equation}

This formulation jointly optimizes battery lifetime and latency compliance. The first term encourages policies that extend device lifetime, while the penalty term discourages configurations that exceed $\tau_{MTP}$. The parameter $\lambda$ controls the relative importance of latency compliance versus energy efficiency. This formulation reflects the practical reality that deterministic latency guarantees are difficult to maintain under stochastic network conditions when computation is offloaded to the edge.
At each decision step, we denote the observable system state as:
\begin{equation}
\mathbf{s}(t) =
[\text{SoC}(t), P_{\text{client}}(t), \text{RTT}(t), B(t), \text{MTP}(t)],
\label{eq:observation_space}
\end{equation}
which captures the information required to assess both energy status and offloading feasibility.

The optimization problem involves several coupled trade-offs. Increasing workload quality improves tracking fidelity but increases processing time and communication overhead, thereby raising both power consumption and latency. Offloading execution reduces local computation but introduces network-dependent delays that may violate latency requirements under adverse conditions. Moreover, execution decisions must adapt over time as battery charge depletes and network conditions fluctuate.

Because future network dynamics and workload evolution are not known apriori, the optimal configuration cannot be determined statically. Instead, effective solutions require online adaptation based on observed system behavior. 

\section{DRL-driven Adaptive Execution Control}
\label{sec:rl_model}

To address the online optimization problem described in the previous section under stochastic network conditions and evolving battery state, we adopt a DRL-driven execution control approach that adapts configuration decisions at runtime based on observed system behavior. Since the problem involves making repeated decisions over time while balancing battery usage and stringent latency requirements under changing network conditions, simple rule-based or threshold policies struggle to handle these trade-offs reliably. On the other hand, a DRL-based controller can adjust decisions online and account for their long-term impact on both latency and energy.

\subsection{DRL-driven Problem Translation}

We formulate the execution management problem as a Markov Decision Process (MDP) defined by the tuple
$\langle \mathcal{S}, \mathcal{A}, \mathcal{P}, \mathcal{R}, \gamma \rangle$,
where the agent makes sequential configuration decisions during system operation.
At each decision epoch $t$, the agent observes the current system state
$\mathbf{s}_t \in \mathcal{S}$ and selects an action
$a_t \in \mathcal{A}$, which determines the execution configuration applied to the XR pipeline.
The system then transitions to a new state $\mathbf{s}_{t+1}$ according to unknown and time-varying dynamics $\mathcal{P}$,
and the agent receives a scalar reward $R_t = \mathcal{R}(\mathbf{s}_t, a_t)$.

The objective of the agent is to learn a policy $\pi: \mathcal{S} \rightarrow \mathcal{A}$
that maximizes the expected discounted return:
\[
\mathbb{E}_{\pi}\!\left[\sum_{k=0}^{\infty} \gamma^{k} R_{t+k+1}\right],
\]
where $\gamma \in (0,1)$ is the discount factor.
This formulation captures the sequential nature of execution decisions,
the stochastic evolution of network conditions and system load,
and the long-term impact of configuration choices on battery depletion.

Unlike offline optimization or static policies, the agent operates \emph{online} and learns directly from observed system behavior during runtime.
The reward function is designed to reflect the dual objectives of maintaining real-time responsiveness
and improving energy efficiency, with particular emphasis on discouraging violations of the MTP latency requirement.
In this way, the MDP formulation provides a principled foundation for adaptive execution control
in latency-sensitive, energy-limited XR systems.

\subsection{State Representation}

The state representation captures the system variables required to assess execution feasibility, energy efficiency, and progress toward battery depletion.
At each decision epoch $t$, the agent observes a state vector
\[
\mathbf{s}_t = [\text{SoC}(t), P_{\text{client}}(t), \text{RTT}(t), B(t), \text{MTP}(t)],
\]
which summarizes the current operating conditions of the XR system.

The battery state of charge $\text{SoC}(t)$ reflects the remaining energy budget and informs how aggressively the controller must conserve power to prolong device lifetime.
The instantaneous client power consumption $P_{\text{client}}(t)$ provides feedback on the energy cost of recent execution decisions and directly relates to the battery depletion dynamics described in Section~\ref{sec:system_model}.
Network round-trip time $\text{RTT}(t)$ and available bandwidth $B(t)$ capture the time-varying communication conditions that influence both the feasibility and latency impact of offloaded execution.
Finally, the measured MTP latency $\text{MTP}(t)$ provides direct feedback on whether recent configurations satisfy real-time responsiveness requirements.

All state variables are normalized to comparable ranges prior to being provided to the learning agent.
This normalization improves numerical stability during training and prevents any single state dimension from dominating the learning process.
Together, these state components enable the agent to reason jointly about latency constraints, energy consumption, network variability, and battery dynamics when selecting execution configurations.

\subsection{Action Space and Control Knobs}

At each decision epoch, the agent selects an execution configuration that determines how the XR workload is processed.
We define a discrete action space
\[
a_t = (q_t, r_t, m_t),
\]
where each action specifies a combination of workload quality, sensing rate, and execution mode.

The quality parameter $q_t$ controls the resolution or fidelity at which visual data is processed, influencing both computational load and data size.
The sensing rate parameter $r_t$ adjusts the sampling frequency of inertial measurements, which affects processing complexity and latency sensitivity.
The execution mode $m_t$ determines whether processing is performed locally on the device or offloaded to a nearby execution resource.
Together, these control knobs capture the primary levers through which execution decisions impact latency, energy consumption, and battery depletion.

The action space is discrete and finite, corresponding to a predefined set of feasible execution configurations.
This design enables fast inference and stable online learning, while still allowing the agent to explore a diverse range of trade-offs between performance and energy efficiency.
By selecting actions from this configuration set at runtime, the agent adapts execution behavior in response to changing network conditions, system load, and battery state.

\subsection{Reward Design and Latency-Prioritized Optimization}

The reward function is designed to guide the learning agent toward execution configurations that maintain real-time responsiveness while improving energy efficiency and battery longevity. 
The function implements the penalized objective in Eq.~(\ref{eq:optimization}), assigning negative reward when MTP exceeds the target threshold and small positive reinforcement when the latency target is met.

At each decision epoch, the agent receives a scalar reward:
\[
R_t = R_{\text{mtp}}(t) + R_{\text{power}}(t) + R_{\text{battery}}(t),
\]
where each term reflects a different system objective.

The primary component, $R_{\text{mtp}}(t)$, captures the effect of MTP latency on system performance. 
This term penalizes configurations that cause the measured MTP latency to exceed the acceptable threshold, with the penalty increasing as the violation magnitude grows. 
Configurations that remain within the latency target receive a small positive reward, reinforcing policies that maintain high latency compliance.

The secondary component, $R_{\text{power}}(t)$, penalizes higher instantaneous client power consumption. 
This term encourages the agent to prefer execution configurations that reduce energy usage when multiple latency-compliant options are available. 
The third component, $R_{\text{battery}}(t)$ provides a modest incentive proportional to the remaining battery state, biasing the policy toward extending operational lifetime as energy resources diminish.

Together, these reward components impose a hierarchical preference structure in which latency compliance is prioritized over power efficiency, and power efficiency is prioritized over long-term battery optimization. 
Rather than enforcing deterministic latency guarantees, this formulation embeds responsiveness requirements directly into the learning objective, enabling the agent to learn policies that minimize latency violations while adapting to stochastic network conditions.

\subsection{DRL Policy Optimization}

To solve the decision problem defined above, we adopt a value-based deep reinforcement learning approach.
The agent learns a policy that maps observed system states to execution configurations by estimating the long-term utility of each possible action under stochastic system dynamics.

We model the action–value function using a Deep Q-Network (DQN), which approximates the optimal action–value function:
\[
Q^*(s,a) = \mathbb{E}\!\left[\sum_{k=0}^{\infty} \gamma^k R_{t+k+1} \,\middle|\, s_t = s, a_t = a \right],
\]
where $s$ is the current system state, $a$ is an execution configuration, $R_t$ is the immediate reward, and $\gamma \in (0,1)$ is a discount factor that controls the trade-off between short-term and long-term objectives.
The choice of DQN is motivated by the fact that the system state comprises multiple real-valued variables, 
that evolve continuously with device and network dynamics. Discretizing these variables for tabular Q-learning would produce a large state space that cannot be adequately explored within the limited number of decisions available during a single operational session. The DQN instead learns a compact function approximation over the continuous state space, allowing generalization across similar states and improving sample efficiency for online learning.

At runtime, the agent selects actions according to the learned Q-function by choosing:
\[
a_t = \arg\max_a Q(s_t, a),
\]
subject to exploration during learning.
This formulation allows the agent to evaluate all feasible execution configurations at each decision point and select the one that maximizes expected cumulative reward.

To stabilize learning under correlated system observations, experience replay is used.
State transitions of the form $(s_t, a_t, R_t, s_{t+1})$ are stored in a replay buffer and randomly sampled during training.
This decorrelates updates and improves sample efficiency.
In addition, a separate target network is maintained to compute stable temporal-difference targets, reducing oscillations during training.

The DQN parameters are optimized by minimizing the mean squared temporal-difference error:
\[
\mathcal{L}(\theta) =
\mathbb{E}\!\left[
\left(
R_t + \gamma \max_{a'} Q(s_{t+1}, a'; \theta^{-})
- Q(s_t, a_t; \theta)
\right)^2
\right],
\]
where $\theta$ denotes the parameters of the online network and $\theta^{-}$ denotes the parameters of the target network.

This DRL formulation enables online adaptation to time-varying network conditions, evolving battery state, and workload dynamics.
By directly learning from system feedback rather than relying on fixed heuristics or offline models, the policy continuously improves its execution decisions while respecting real-time latency sensitivity.

\subsection{Exploration Strategy}

To balance exploration and exploitation during online learning, the agent adopts an $\epsilon$-greedy decision policy. At each decision step $t$, the agent selects a random execution configuration with probability $\epsilon_t$, and otherwise selects the configuration with the highest estimated action-value under the current Q-function:
\begin{equation}
a_t =
\begin{cases}
\text{random action from } \mathcal{A}, & \text{with probability } \epsilon_t, \\
\operatorname*{arg\,max}\limits_{a \in \mathcal{A}} Q(s_t,a), & \text{with probability } 1-\epsilon_t.
\end{cases}
\label{eq:epsilon_greedy}
\end{equation}

where $\mathcal{A}$ denotes the discrete set of execution configurations.

The exploration rate $\epsilon_t$ is decayed over time, allowing the agent to initially explore a wide range of configurations under varying system and network conditions, and gradually shift toward stable exploitation of configurations that satisfy latency requirements while reducing power consumption. Exploration operates over complete execution configurations, ensuring that learning respects the discrete structure of the control space.

\subsection{Integration and Decision Frequency}

\begin{figure}
    \centering
    \includegraphics[width=1\linewidth]{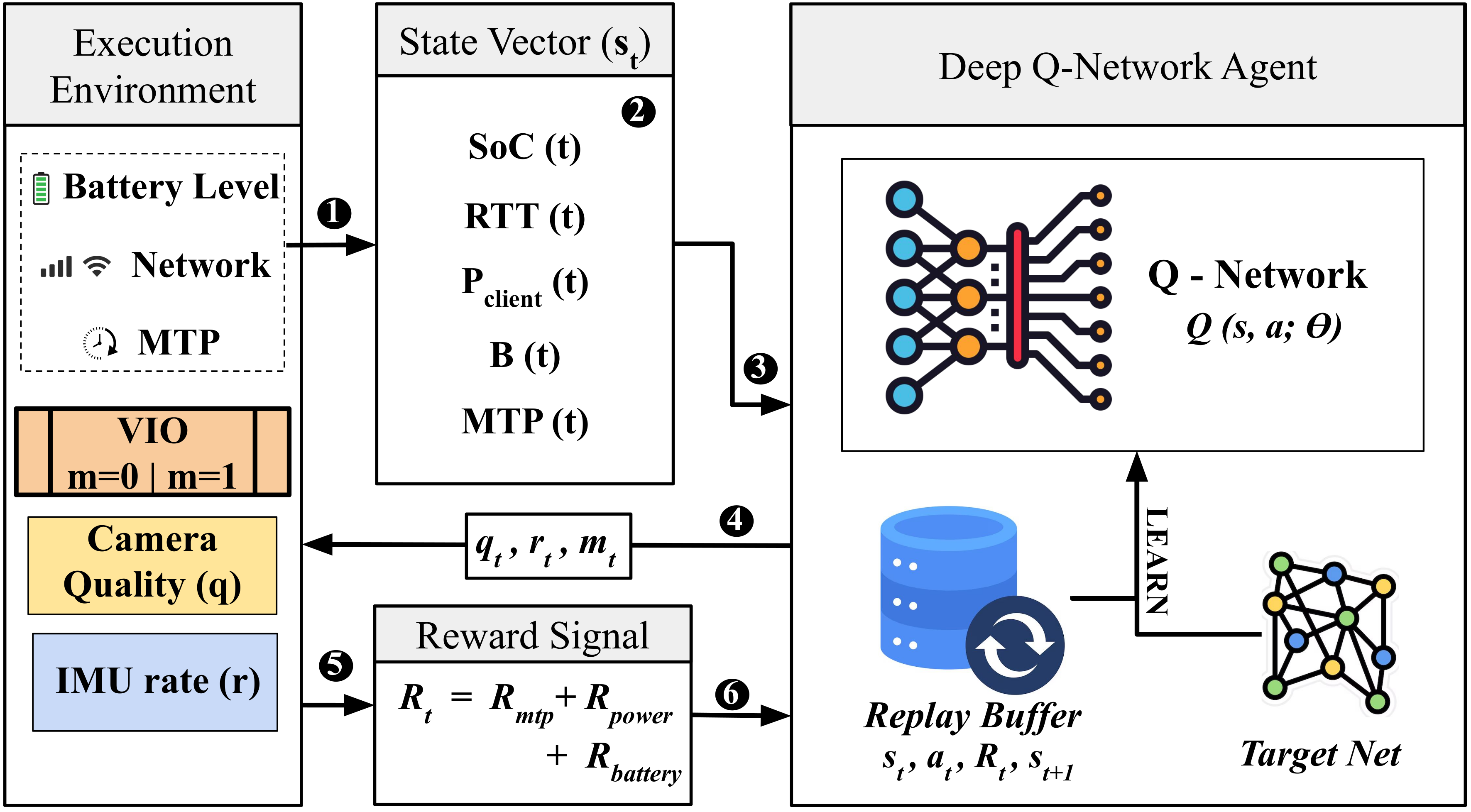}
    \caption{Reinforcement learning control architecture where the DQN agent continuously observes battery, network, and latency conditions to select optimal quality, IMU rate, and processing mode configurations that extend device lifetime under real-time constraints.}
    \label{fig:rl_model}
    \vspace{-0.2in}
\end{figure}

The learning-based controller runs as a periodic control loop that operates independently from the latency-sensitive XR pipeline. Instead of making decisions for every frame or pose update, the controller observes system state continuously but updates execution configurations at a coarse time scale.

In our design, the controller evaluates system state regularly but selects a new execution configuration once per second. Each selected configuration remains in effect until the next decision point. This coarse-grained control avoids frequent reconfiguration while still allowing the system to adapt to changes in battery state and network conditions.

This design ensures that learning and decision-making do not execute on the MTP sensitive path. The controller runs in parallel with sensing, tracking, and rendering components and applies configuration updates asynchronously. As a result, the XR pipeline continues to operate without interruption while the controller adapts execution parameters over time. Fig. \ref{fig:rl_model} illustrates the complete learning and control loop that operationalizes the system model introduced in Fig. \ref{fig:setup}. Runtime signals from the execution environment (client in Fig.~\ref{fig:setup}), including battery state, network conditions, and measured MTP latency, are continuously monitored and aggregated into a state vector that represents the current operating condition of the XR system. At each decision interval, the DQN-based controller consumes this state and selects an execution configuration $(q, r, m)$, corresponding to the camera resolution, IMU rate, and execution mode. This configuration is applied back to the VIO pipeline, influencing subsequent execution behavior and system measurements. The resulting latency and power outcomes are then used to compute a reward signal, which closes the loop by updating the learning agent without interfering with the latency-sensitive execution path.

\section{Evaluation and Results}
\label{sec:eval}
In this section, we describe the implementation details of the proposed framework, followed by the experimental configuration used to evaluate the system. We then present the results of this evaluation.

\subsection{Implementation}
We implement the proposed framework in an end-to-end XR runtime using \textsc{ILLIXR}~\cite{illixr}, a modular, open-source XR systems research platform that enables closed-loop, system-level evaluation under near realistic MTP latency requirements. Both the headset client and the edge server are implemented as ILLIXR runtimes and deployed on two separate machines, each equipped with an 8-core CPU, 32 GB RAM, and
an NVIDIA RTX 2000 Ada GPU. 
The implementation supports dynamic selection among discrete execution configurations, described in Section \ref{sec:execution-mode}, at runtime, based on observed battery state, power proxies, MTP latency, and network conditions. The lightweight DRL controller operates online, without offline training, and is designed as a plugin to run asynchronously with the XR pipeline. This implementation enables direct comparison of local and edge-assisted execution strategies under identical runtime conditions.

\subsection{Experimental Configuration}
We evaluate the proposed framework under a controlled setup with fixed execution options and time-varying network conditions. Execution decisions are applied periodically based on observed system state. MTP latency is treated as a latency-sensitive performance objective and evaluated relative to an application-level threshold. While immersive XR systems typically target an MTP budget of around 20 ms, we use a slightly relaxed threshold of 30 ms to account for experimental overhead. Battery behavior is simulated via a runtime plugin that updates state-of-charge based on configuration-dependent power consumption. Network conditions are varied using bandwidth profiles spanning favorable and constrained regimes. Across all experiments, we compare our proposed adaptive policy under identical workloads and network traces with static policies as well as adaptive policies. Experimental parameters are summarized in Tab.~\ref{tab:exp_params}.
To account for the stochastic nature of online reinforcement learning, we repeat each experiment three times with independent random seeds. For DRL configurations, each run trains a fresh DQN agent from a different random initialization, producing independent learned policies and exploration trajectories. 

\begin{table}[t]
\centering
\caption{Experimental configuration parameters}
\label{tab:exp_params}

\setlength{\tabcolsep}{3pt}
\renewcommand{\arraystretch}{1.2}

\begin{tabular}{|l|c|}
\hline
{\fontsize{8}{9}\selectfont\textbf{Category}} &
{\fontsize{8}{9}\selectfont\textbf{Value}} \\
\hline

\multicolumn{2}{|l|}{\scriptsize\textbf{Control and Timing}} \\
\hline
\scriptsize Decision interval & \scriptsize 1 s \\
\hline
\scriptsize Target frame duration & \scriptsize 50 ms (20 Hz) \\
\hline
\scriptsize MTP latency threshold & \scriptsize 30 ms \\
\hline

\multicolumn{2}{|l|}{\scriptsize\textbf{Battery Model}} \\
\hline
\scriptsize Simulated battery capacity & \scriptsize 16.6 Wh \\
\hline
\scriptsize Initial state-of-charge & \scriptsize 100\% \\
\hline
\scriptsize Baseline power & \scriptsize 0.5 W \\
\hline

\multicolumn{2}{|l|}{\scriptsize\textbf{Execution Configurations}} \\
\hline
\scriptsize Image quality levels ($q$) & \scriptsize $\{376\times240,\ 564\times360,\ 752\times480\}$ \\
\hline
\scriptsize IMU sampling rates ($r$) & \scriptsize $\{200,\ 150,\ 100\}$ Hz \\
\hline
\scriptsize Execution modes ($m$) & \scriptsize \{LOCAL, OFFLOAD\} \\
\hline
\scriptsize Total configurations & \scriptsize 18 \\
\hline

\multicolumn{2}{|l|}{\scriptsize\textbf{Network Conditions}} \\
\hline
\scriptsize Bandwidth profiles & \scriptsize $\{1,\ 10,\ 100,\ 500,\ 1000\}$ Mbps \\
\hline
\scriptsize Profile duration & \scriptsize 60 s \\
\hline
\scriptsize Bandwidth cycle length & \scriptsize 300 s \\
\hline

\end{tabular}

\vspace{-0.2in}
\end{table}

\subsection{Results}

We organize the results to progressively examine different aspects of system behavior. We first study energy consumption and MTP latency under stable network conditions to establish baseline trade-offs between local, offloaded, and RL-Adaptive execution. We then evaluate performance under bandwidth variability and network stress. We also examine the policies learned by the RL-Adaptive method over the different network conditions. We do similar evaluations for certain baseline adaptive strategies and their comparison with RL-Adaptive, studying the nuanced interpretation of the learned policies over performance. This is followed by a hyperparameter analysis that isolates the impact of learning hyperparameters on latency compliance and execution-mode selection. The evaluation ends with the analysis of the overhead that the DRL plugin employs over the static system.
\begin{figure}[htbp]
    \centering

    \begin{subfigure}{0.48\linewidth}
        \centering
        \includegraphics[width=\linewidth]{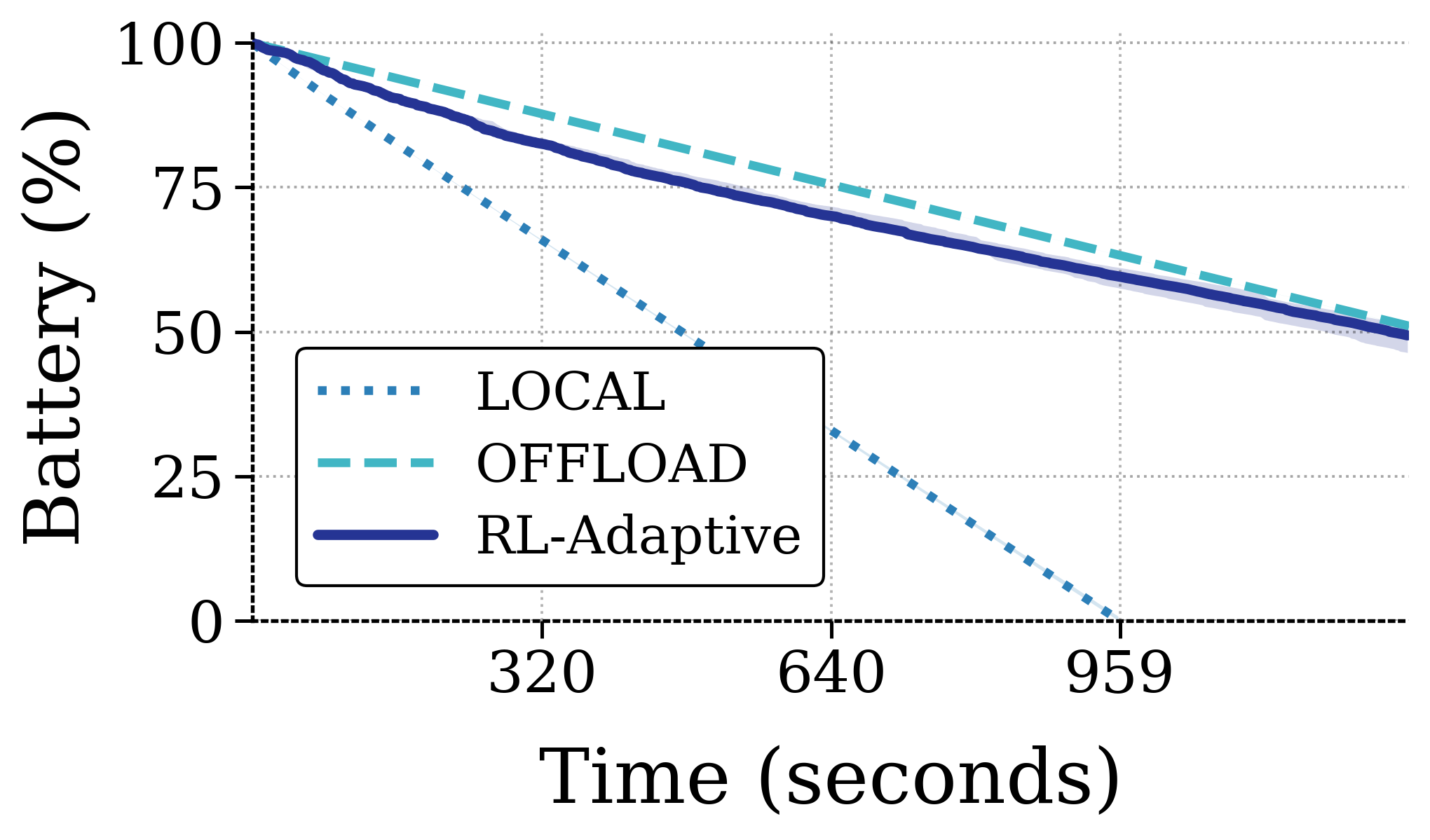}
        \caption{Battery depletion over time}
        \label{fig:result1_energy-a}
    \end{subfigure}
    \hspace{0.005in}
    \begin{subfigure}{0.48\linewidth}
        \centering
        \includegraphics[width=\linewidth]{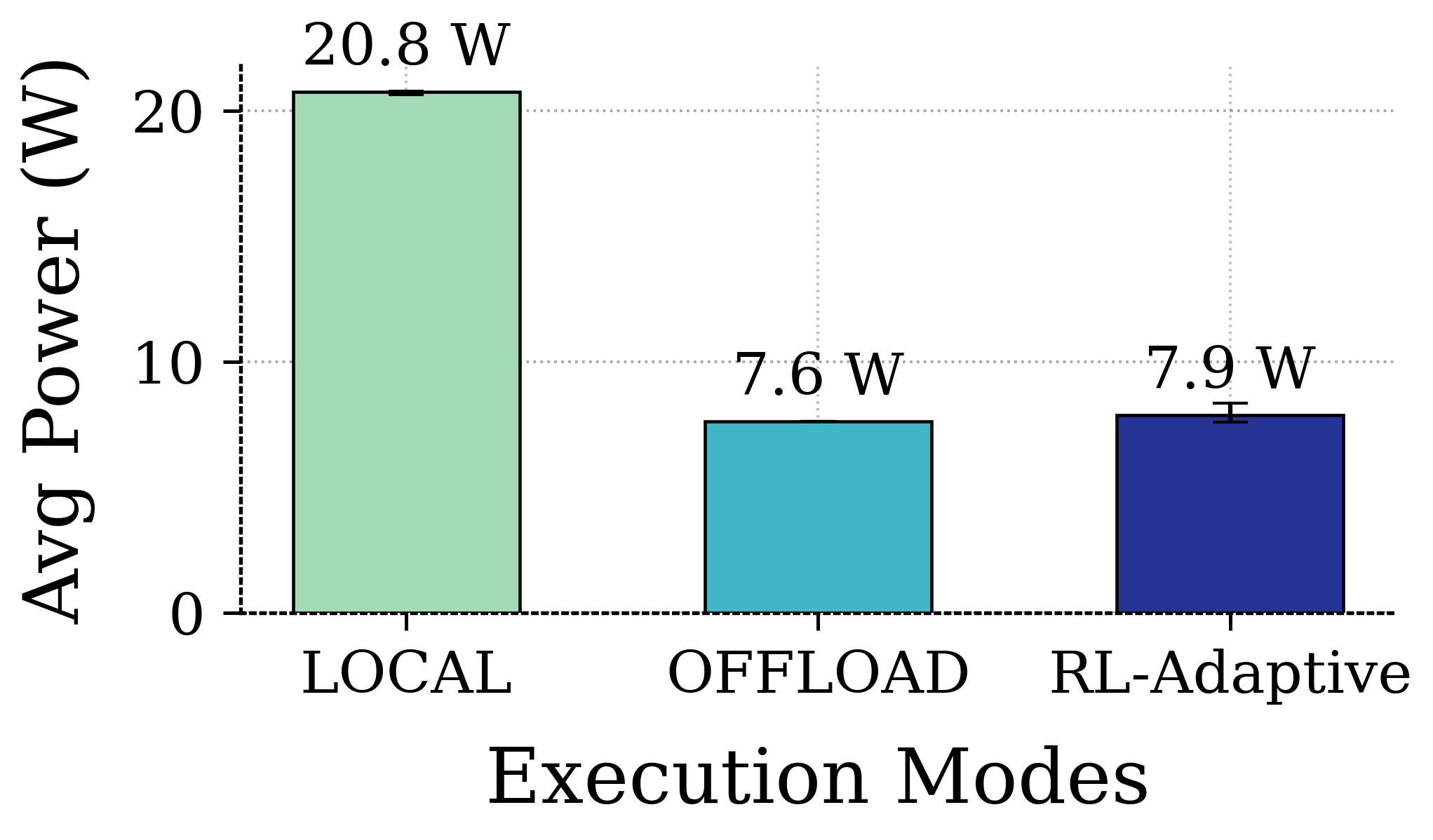}
        \caption{Average power consumption}
        \label{fig:result1_energy-b}
    \end{subfigure}

    \caption{Energy characteristics of LOCAL, OFFLOAD, and RL-Adaptive execution strategies under stable network condition.}
    \label{fig:result1_energy}
    \vspace{-0.2in}
\end{figure}

\subsubsection{Performance under Stable Network Conditions}
We first compare LOCAL, OFFLOAD, and RL-Adaptive execution under stable network conditions. Fig. \ref{fig:result1_energy-a} shows battery depletion over time, while Fig. \ref{fig:result1_energy-b} reports average power consumption. LOCAL execution exhibits the highest power draw (20.8 W), rapidly depleting the battery and terminating execution after approximately 960 s. OFFLOAD minimizes power consumption (7.6 W), reducing energy use by 63.3\% relative to LOCAL and preserving more than half of the battery capacity over the same duration. 
The RL-Adaptive policy achieves near-OFFLOAD power consumption, averaging 7.9 W across seeds and extending projected battery lifetime by approximately 163\% compared to LOCAL (42 vs. 16 minutes) while completing the full evaluation window. Notably, the RL policy achieves near OFFLOAD like battery lifetime. RL-Adaptive's battery drain curve closely tracks the OFFLOAD trajectory, indicating that the learned policy achieves comparable energy efficiency to full offloading.

\begin{figure}[t]
\centering

\begin{subfigure}{0.45\linewidth}
    \centering
    \includegraphics[width=\linewidth]{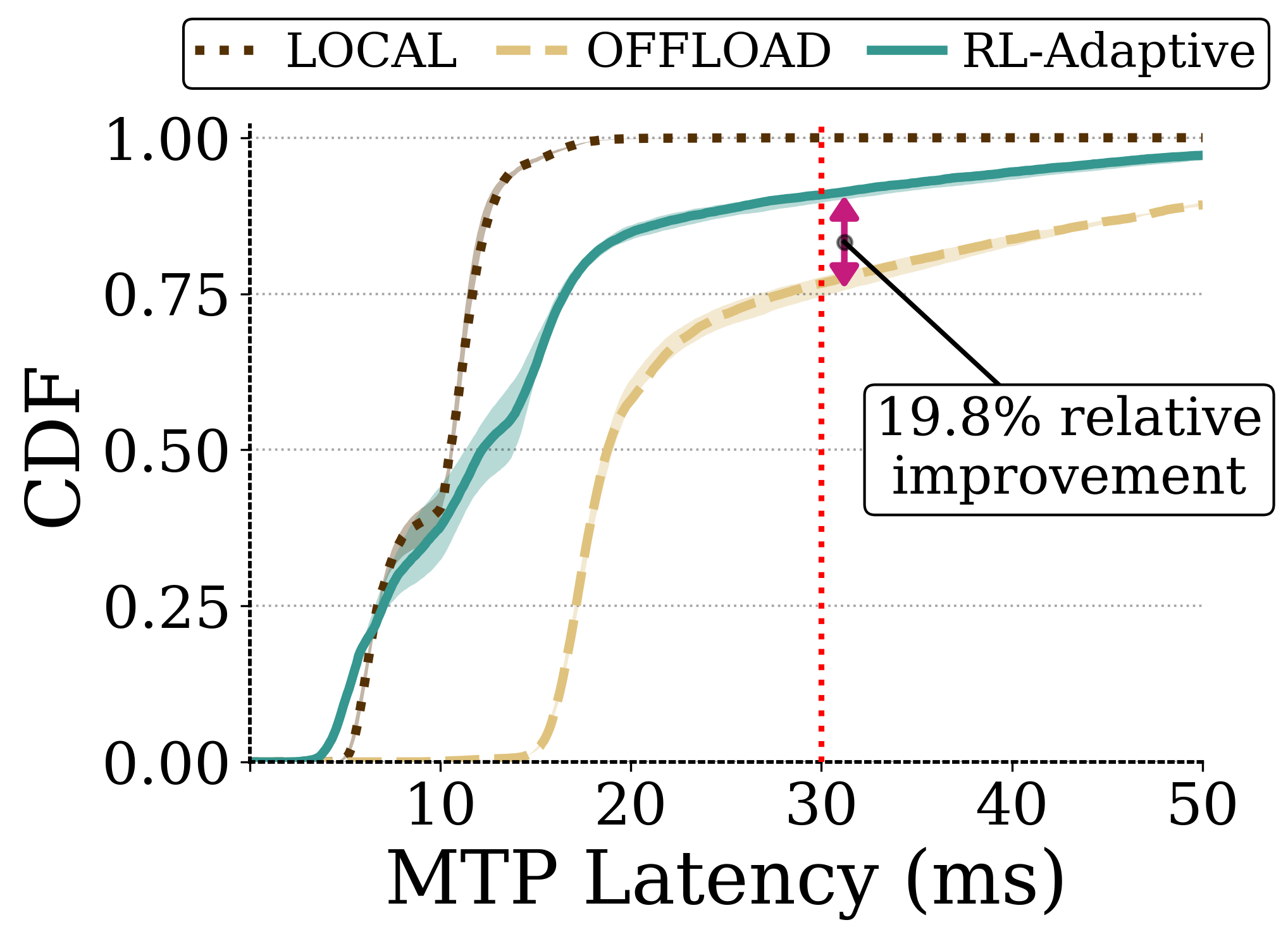}
    \caption{CDF of MTP latency distribution under different execution strategies under normal network condition.}
    \label{fig:result1_mtp}
\end{subfigure}
\begin{subfigure}{0.53\linewidth}
    \centering
    \includegraphics[width=\linewidth]{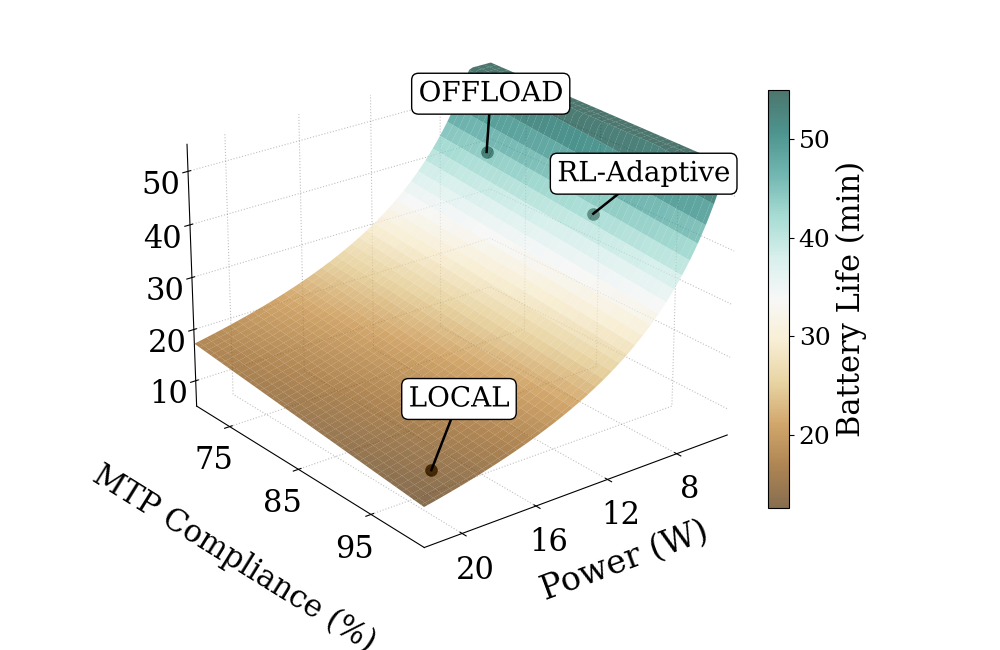}
    \caption{Multi-objective trade-off between power consumption, MTP compliance, and battery life under stable network condition.}
    \label{fig:result1_pareto}
\end{subfigure}

\caption{Performance characteristics of different execution strategies under stable network conditions, showing the trade-offs between energy efficiency and MTP latency behavior.}

\vspace{-0.2in}
\end{figure}

Fig. \ref{fig:result1_mtp} reports the cumulative distribution of MTP latency relative to the 30 ms threshold. LOCAL execution maintains 100\% compliance due to the absence of network delay, whereas OFFLOAD exhibits substantial variability, with only 75.8\% of frames meeting the latency requirement. In contrast, the RL-Adaptive policy improves compliance to 90.8\%, demonstrating the benefit of dynamically adapting execution configurations in response to system conditions.
The shaded envelope for RL-Adaptive reflects variance across three independently trained policies. Despite different random initializations, all seeds achieve compliance between 89.5\% and 91.5\%, demonstrating that the DQN agent reliably learns effective policies from online interaction alone. At the 30 ms threshold, RL improves compliance by 15.0 percentage points over OFFLOAD, corresponding to a 19.8\% relative improvement in the fraction of frames meeting the latency requirement. Importantly, this improvement is achieved while consuming only marginally more power than full offloading, illustrating that adaptive execution can achieve near-offload energy efficiency without sacrificing latency robustness.

Fig. \ref{fig:result1_pareto} shows the trade-off between power, MTP compliance, and predicted battery lifetime. LOCAL achieves perfect compliance at high power, while OFFLOAD minimizes power but reduces compliance. The RL-Adaptive policy achieves 90.8\% compliance at near-OFFLOAD like power, effectively Pareto-dominating OFFLOAD by delivering higher compliance at nearly identical energy cost.

\subsubsection{Power Usage and Efficiency}

\begin{figure}[t]
    \centering

    \begin{subfigure}[b]{0.48\linewidth}
        \centering
        \includegraphics[width=\linewidth]{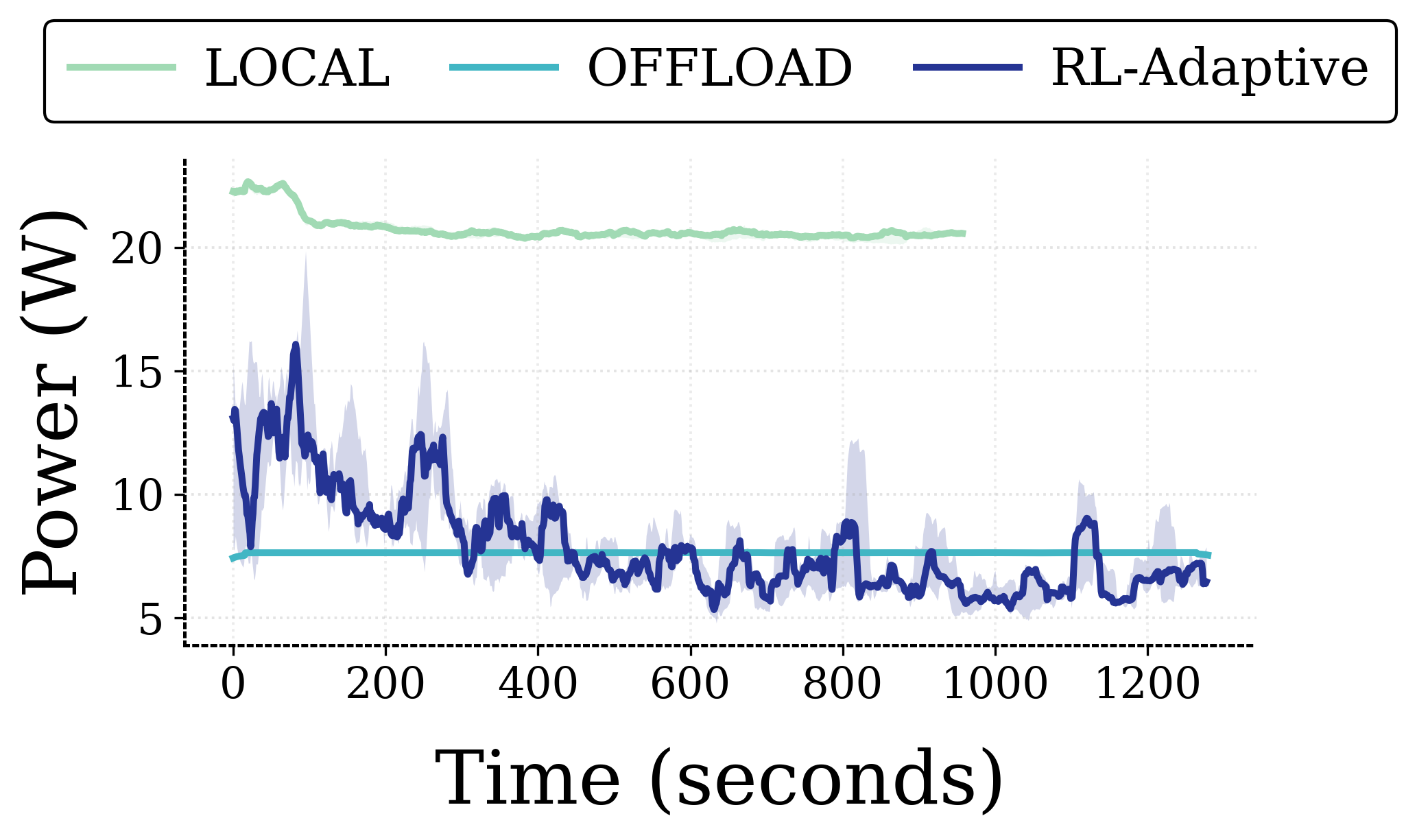}
        \caption{Power consumption over time}
        \label{fig:result2a}
    \end{subfigure}
    \hspace{0.005in}
    \begin{subfigure}[b]{0.48\linewidth}
        \centering
        \includegraphics[width=\linewidth]{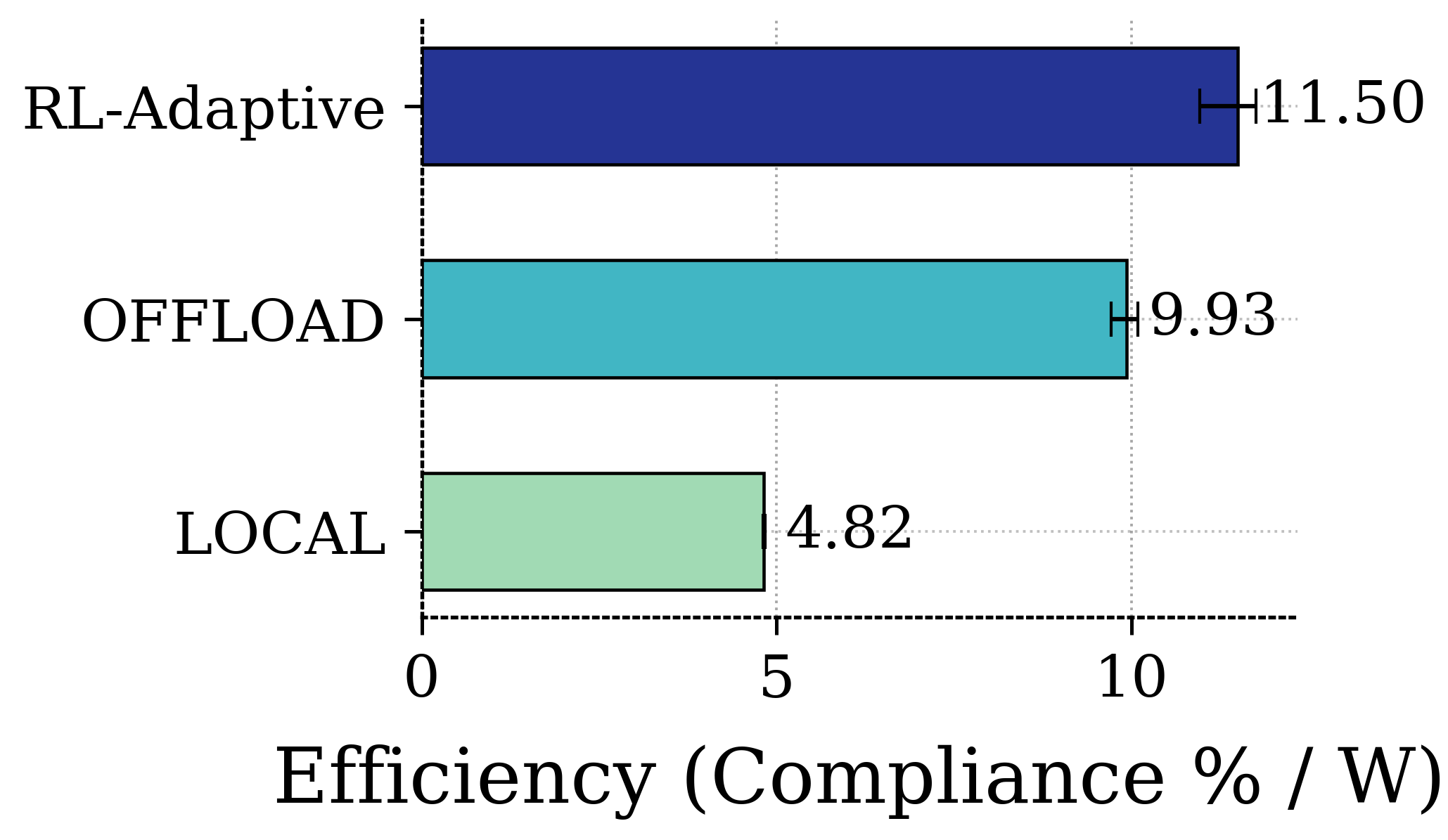}
        \caption{Efficiency Mteric}
        \label{fig:result2b}
    \end{subfigure}

    \caption{Power consumption and efficiency in terms of MTP compliance per watt during normal network condition.}
    \label{fig:result2}
    \vspace{-0.3in}
\end{figure}

Next, we examine the power consumption dynamics underlying the latency–energy trade-off across execution strategies. Fig. \ref{fig:result2a} shows smoothed power traces over time, computed using a rolling average to suppress high-frequency measurement noise and emphasize sustained consumption trends. LOCAL exhibits a consistently high power draw (20.8W) due to continuous on-device processing and depletes the battery after approximately 960 s, preventing completion of the full experiment. OFFLOAD maintains a low and stable power profile (7.6W), enabling completion of the workload with substantial battery remaining. The RL-Adaptive strategy initially exhibits higher power consumption during the exploration phase as the DQN agent evaluates diverse execution configurations, before converging toward a power profile comparable to full offloading. Although short-term fluctuations remain due to adaptive mode switching, the learned policy predominantly selects energy-efficient configurations while maintaining MTP compliance. Overall, proposed RL reduces power consumption by 62.0\% relative to LOCAL (Fig. \ref{fig:result1_energy-b}), nearly matching the reduction achieved by full offloading (63.3\%), illustrating how adaptive execution can achieve OFFLOAD-level energy efficiency without fully committing to OFFLOAD-ing.

Fig. \ref{fig:result2b} analyzes efficiency using a compliance-per-watt metric that relates MTP compliance to average power consumption. LOCAL achieves perfect compliance but at the cost of high power consumption (4.82 compliance/W). OFFLOAD improves efficiency due to its low power consumption (9.93 compliance/W) but fails to meet the target latency requirement, achieving only 75.8\% compliance. In contrast, the RL-Adaptive strategy achieves the highest efficiency among all approaches, reaching 11.50 compliance/W while maintaining 90.8\% compliance at 7.9 W average power. Unlike OFFLOAD, which sacrifices responsiveness for energy savings, the RL-Adaptive policy simultaneously achieves high efficiency and satisfies the latency requirement, demonstrating that \emph{adaptive execution can jointly optimize energy efficiency and real-time responsiveness in latency-sensitive XR systems.}

\begin{figure*}[tb]
    \centering

    \begin{subfigure}{0.27\textwidth}
        \centering
        \includegraphics[width=\linewidth]{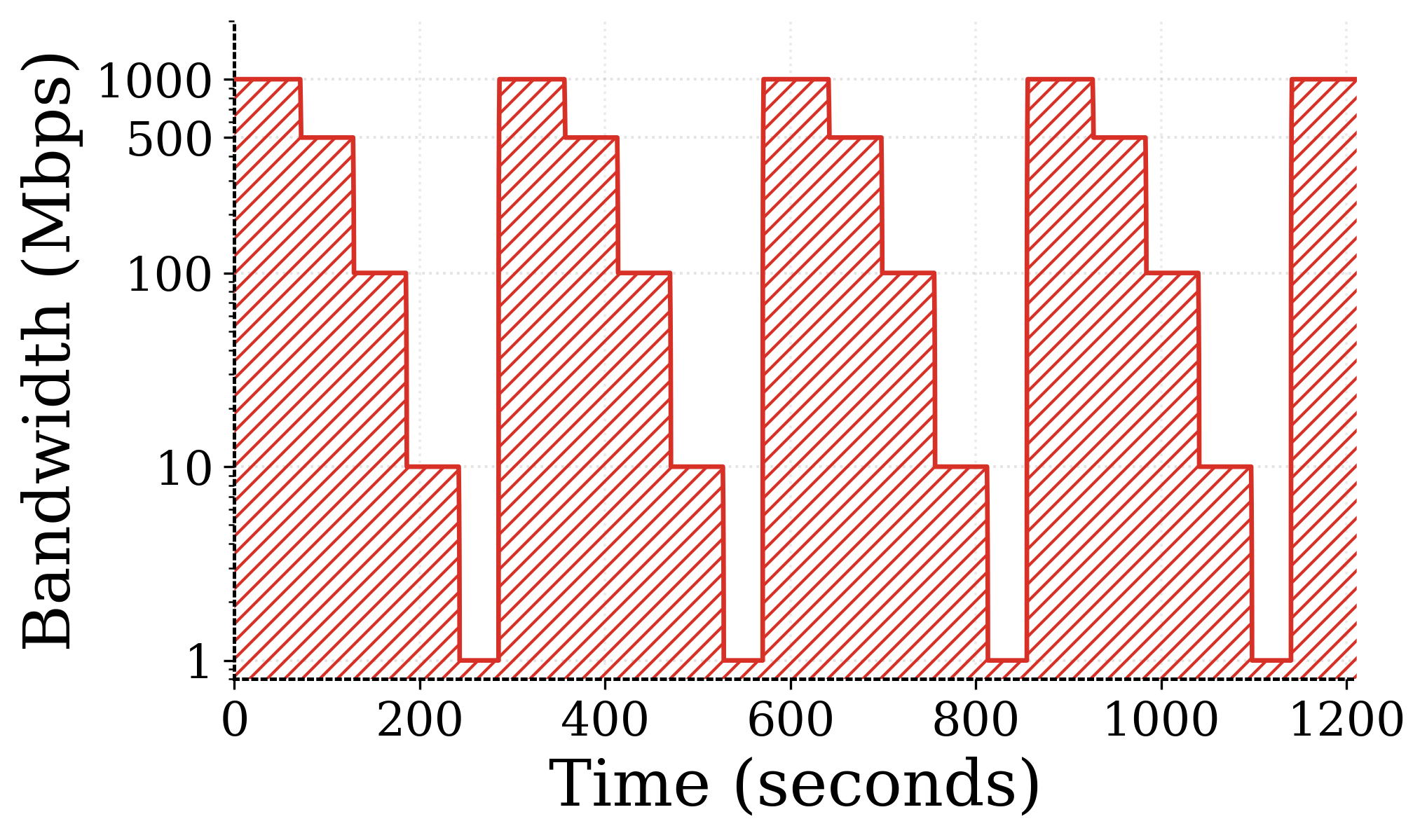}
        \caption{Bandwidth profile for creating network disruptions.}
        \label{fig:result3_triptych-a}
    \end{subfigure}
    \begin{subfigure}{0.27\textwidth}
        \centering
        \includegraphics[width=\linewidth]{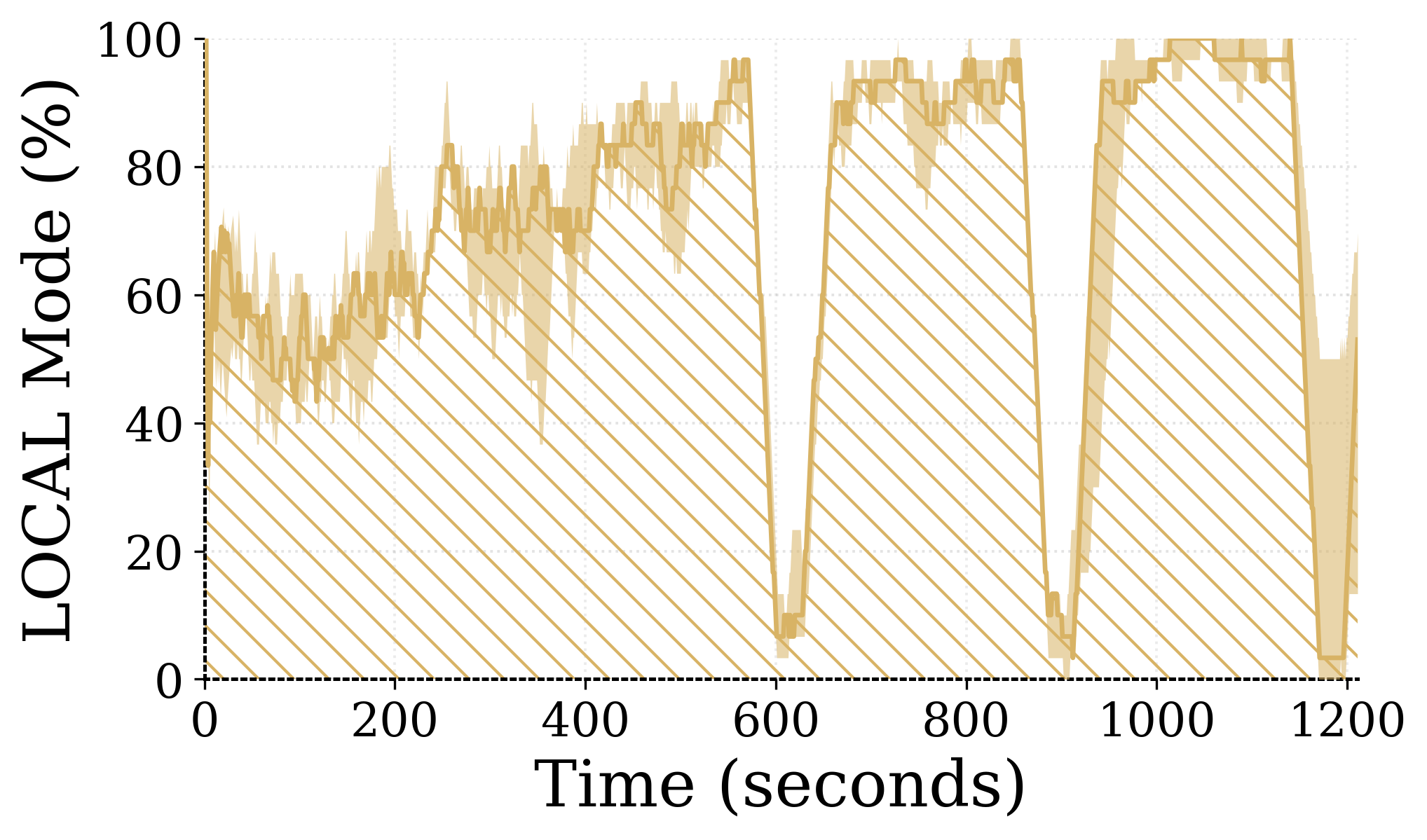}
        \caption{RL mode selection during network disruptions.}
        \label{fig:result3_triptych-b}
    \end{subfigure}
    \begin{subfigure}{0.27\textwidth}
        \centering
        \includegraphics[width=\linewidth]{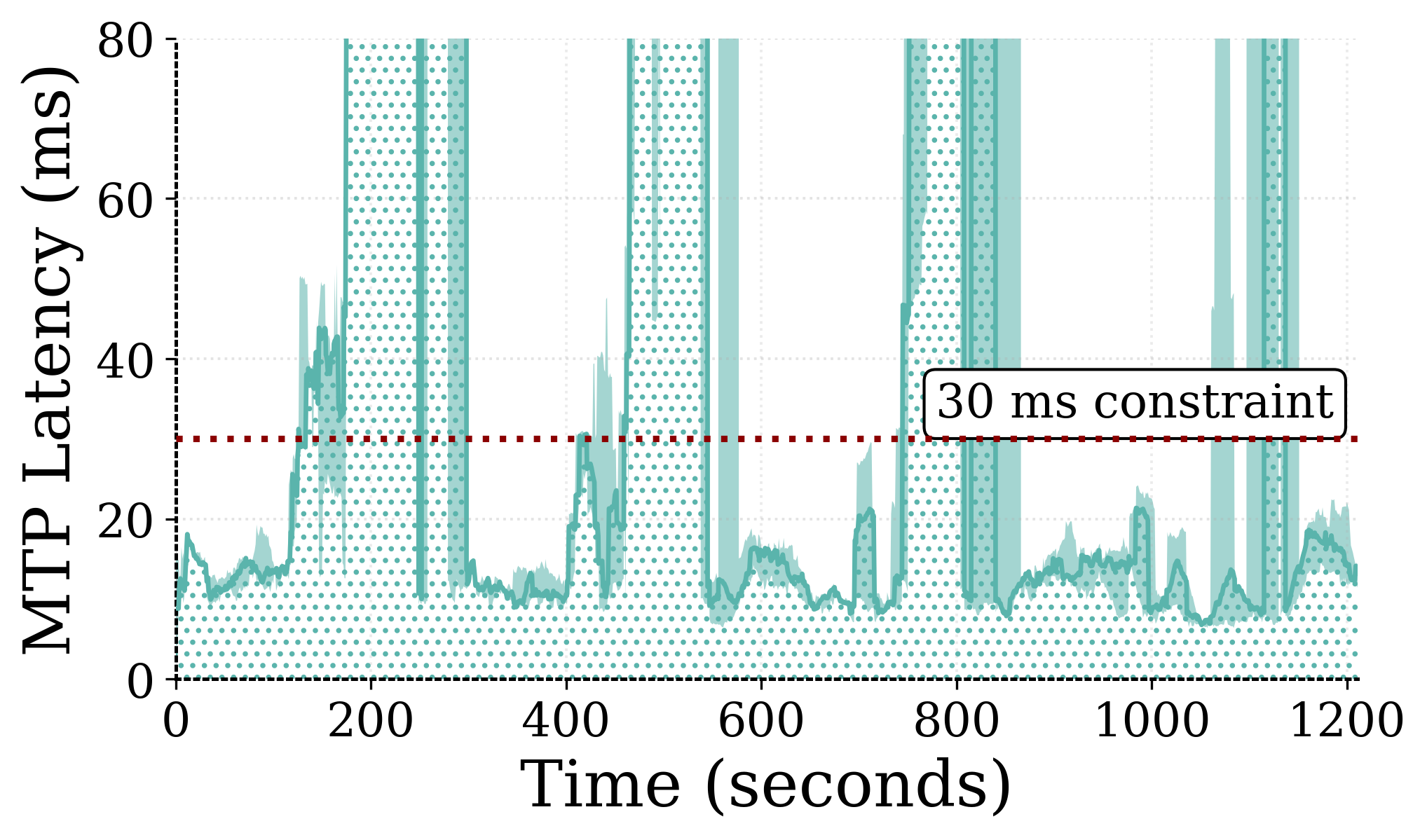}
        \caption{MTP latency over time during network disruptions.}
        \label{fig:result3_triptych-c}
    \end{subfigure}
    \caption{RL behavior under network variability: bandwidth trace (a), resulting MTP latency over time (b), and execution mode switching (c).}
    \label{fig:result3_triptych}
    \vspace{-0.2in}
\end{figure*}

\subsubsection{Performance under Variable Network Conditions}
To evaluate the robustness of execution decisions under realistic network variability, we subject the system to cyclic bandwidth fluctuations spanning three orders of magnitude. The available bandwidth alternates between 1 Gbps, 500 Mbps, 100 Mbps, 10 Mbps, and 1 Mbps, with each level held for 60 s, resulting in a 300 s cycle that repeats approximately four times over the experiment duration. Fig. \ref{fig:result3_triptych-a} shows the resulting bandwidth profile using a log-scaled axis. As in earlier figures, the trace shows the median across three seeds, with the bandwidth profile identical across seeds since it is deterministically generated by the network emulator. To reduce measurement noise while preserving sharp regime transitions, the plotted trace applies peak-preserving smoothing using a rolling maximum with a window of 15 samples.

Figs. \ref{fig:result3_triptych-b} and \ref{fig:result3_triptych-c} illustrate how the RL controller adapts execution decisions under bandwidth variability. The mode-selection trace in Fig. \ref{fig:result3_triptych-b}, computed using a rolling window of 30 decisions, shows that the agent progressively learns to anti-correlate execution mode with available bandwidth. During the initial exploration phase (0–300s) LOCAL usage remains noisy across bandwidth levels. By the second cycle (300–600s) adaptation begins to emerge, and after approximately 600s the policy becomes sharply responsive. LOCAL execution approaches 100\% during 1–10 Mbps phases and drops close to zero during 500–1000 Mbps phases. This improved switching behavior directly reduces latency violations. The MTP trace in Fig. \ref{fig:result3_triptych-c} shows that early bandwidth cycles produce wide and sustained latency spikes, while later cycles contain only brief violations corresponding to the 1s decision interval during bandwidth transitions. The envelope across seeds narrows after convergence, indicating that independently trained agents learn similar policies despite different random initializations.

\begin{figure}[t]
    \centering

    \begin{subfigure}{0.48\linewidth}
        \centering
        \includegraphics[width=\linewidth]{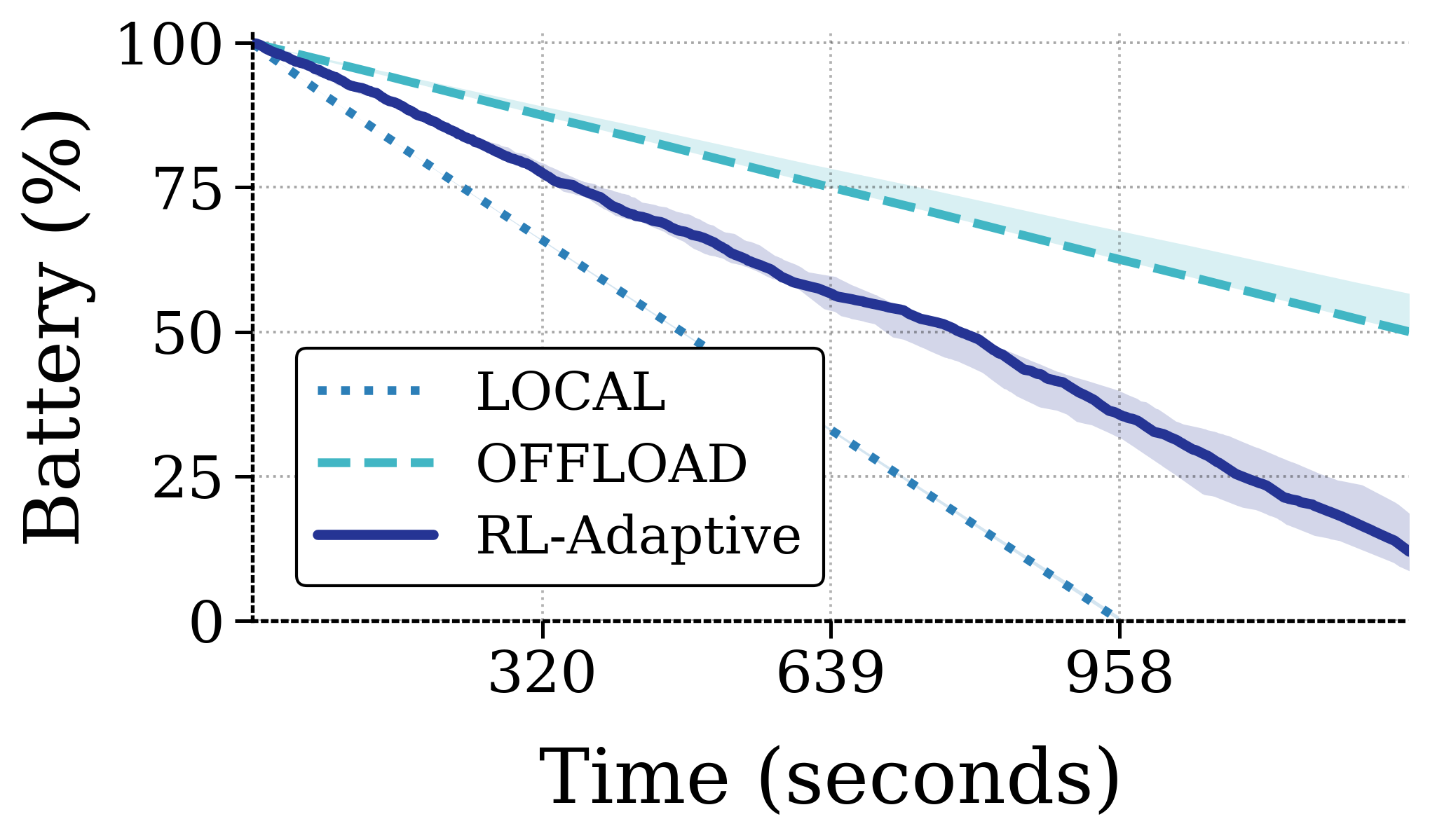}
        \caption{Battery depletion over time}
        \label{fig:result3_energy-a}
    \end{subfigure}
    \hspace{0.005in}
    \begin{subfigure}{0.48\linewidth}
        \centering
        \includegraphics[width=\linewidth]{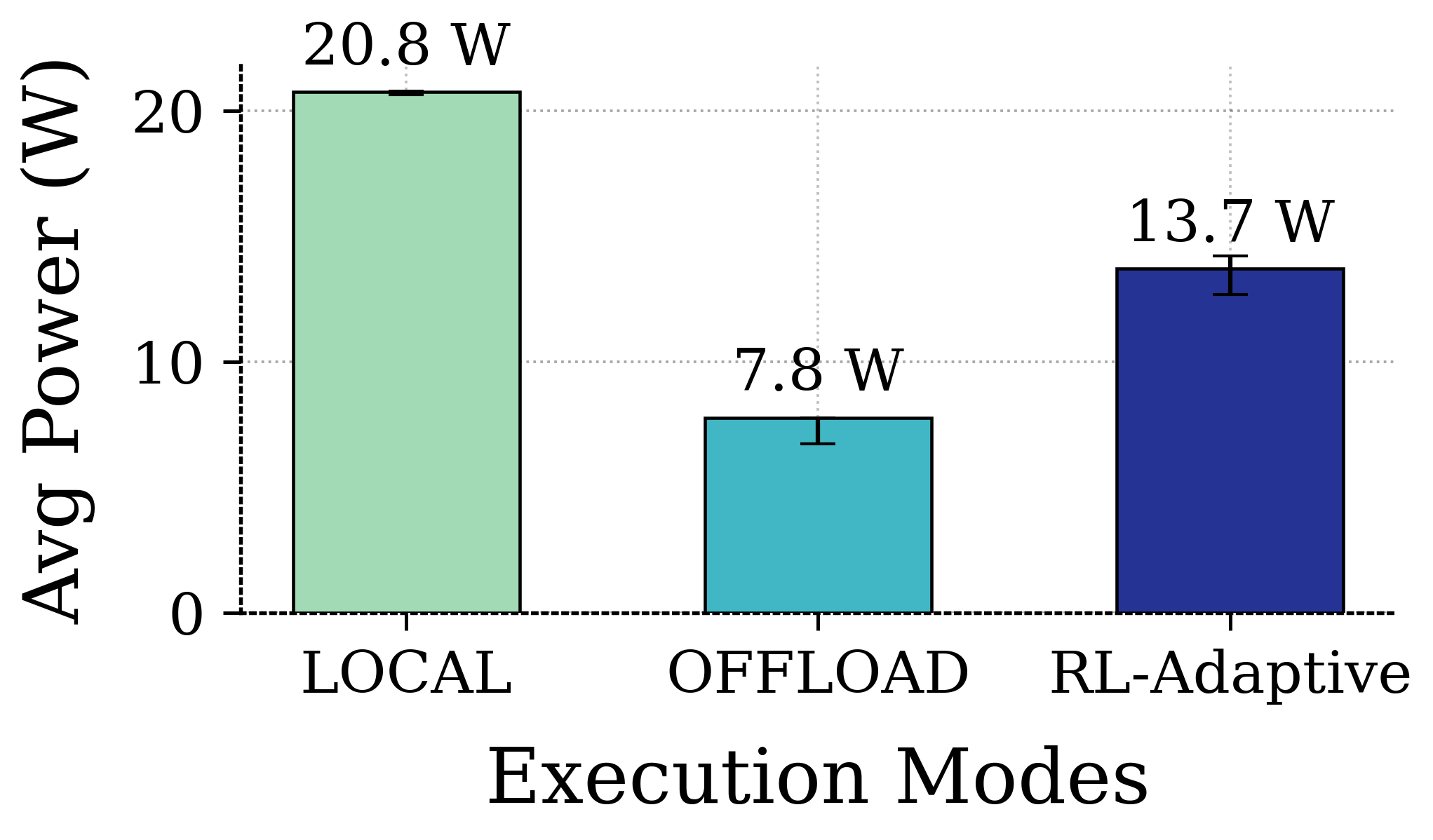}
        \caption{Average power consumption}
        \label{fig:result3_energy-b}
    \end{subfigure}

    \caption{Energy characteristics of LOCAL, OFFLOAD, and RL-Adaptive execution strategies under variable network condition.}
    \label{fig:result3_energy}
    \vspace{-0.2in}
\end{figure}

Fig. \ref{fig:result3_energy-a} shows battery drain under variable bandwidth conditions. LOCAL depletes the battery at approximately 960s, identical to the stable scenario, confirming that local execution is independent of network conditions. OFFLOAD maintains the slowest drain rate, while the RL-Adaptive policy drains faster than OFFLOAD but remains substantially more efficient than LOCAL. This increased drain occurs because the agent executes locally more frequently and uses higher VIO quality settings, when bandwidth becomes unreliable in order to preserve latency compliance. Fig. \ref{fig:result3_energy-b} compares average power consumption under variable network conditions. LOCAL power remains unchanged at 20.8W, while OFFLOAD remains low at 7.8W (similar to 7.6W under stable bandwidth). The RL-Adaptive policy consumes 13.7W, higher than its 7.9W stable-condition power but still 34.1\% lower than LOCAL, reflecting the additional on-device computation required to maintain responsiveness under network variability.

\begin{figure}[htbp]
    \centering

    \begin{subfigure}{0.48\linewidth}
        \centering
        \includegraphics[width=\linewidth]{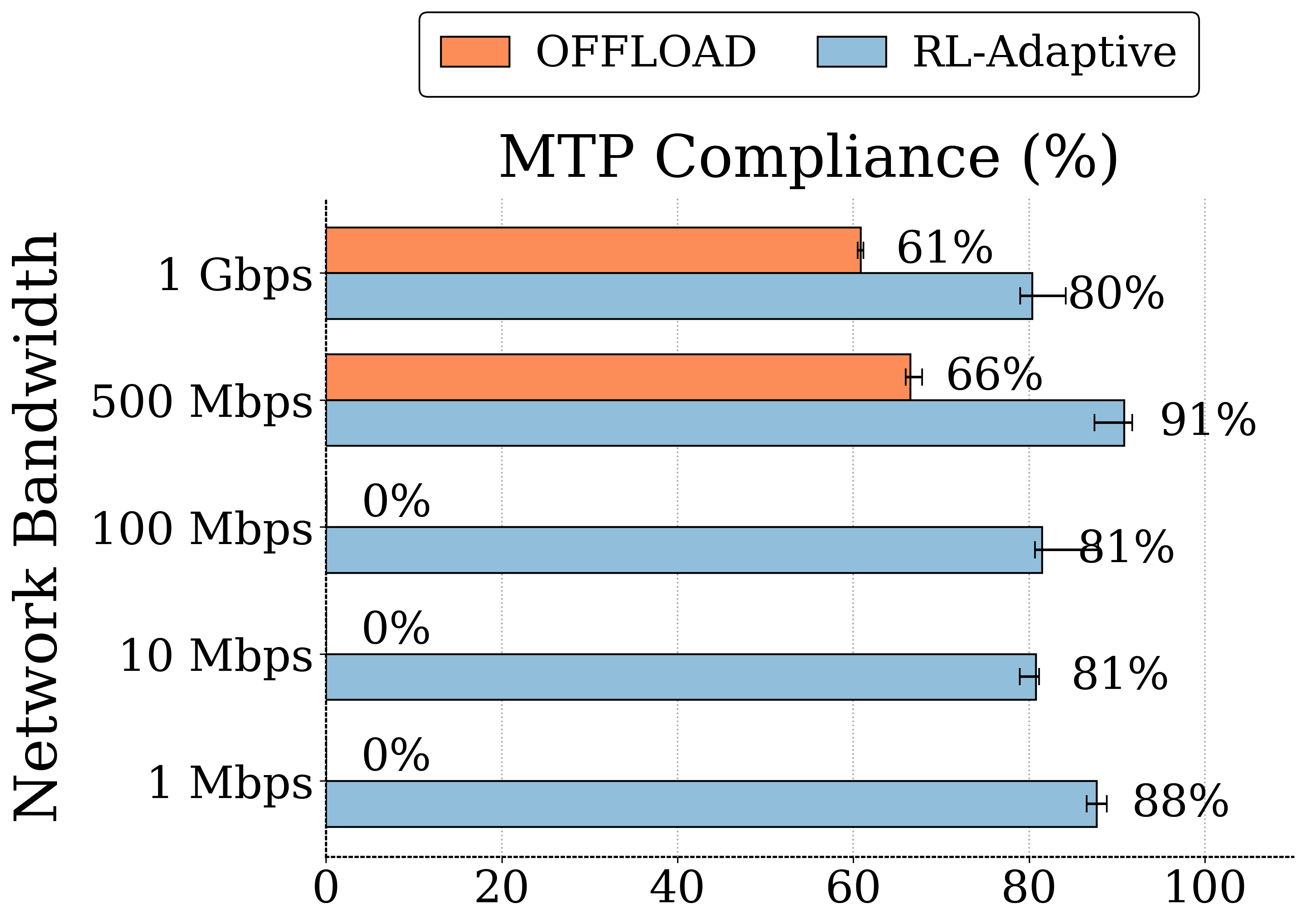}
        \caption{MTP compliance vs. network bandwidth (OFFLOAD vs. RL-Adaptive).}
        \label{fig:result3_compliance_bar}
    \end{subfigure}
    \hspace{0.005in}
    \begin{subfigure}{0.48\linewidth}
        \centering
        \includegraphics[width=\linewidth]{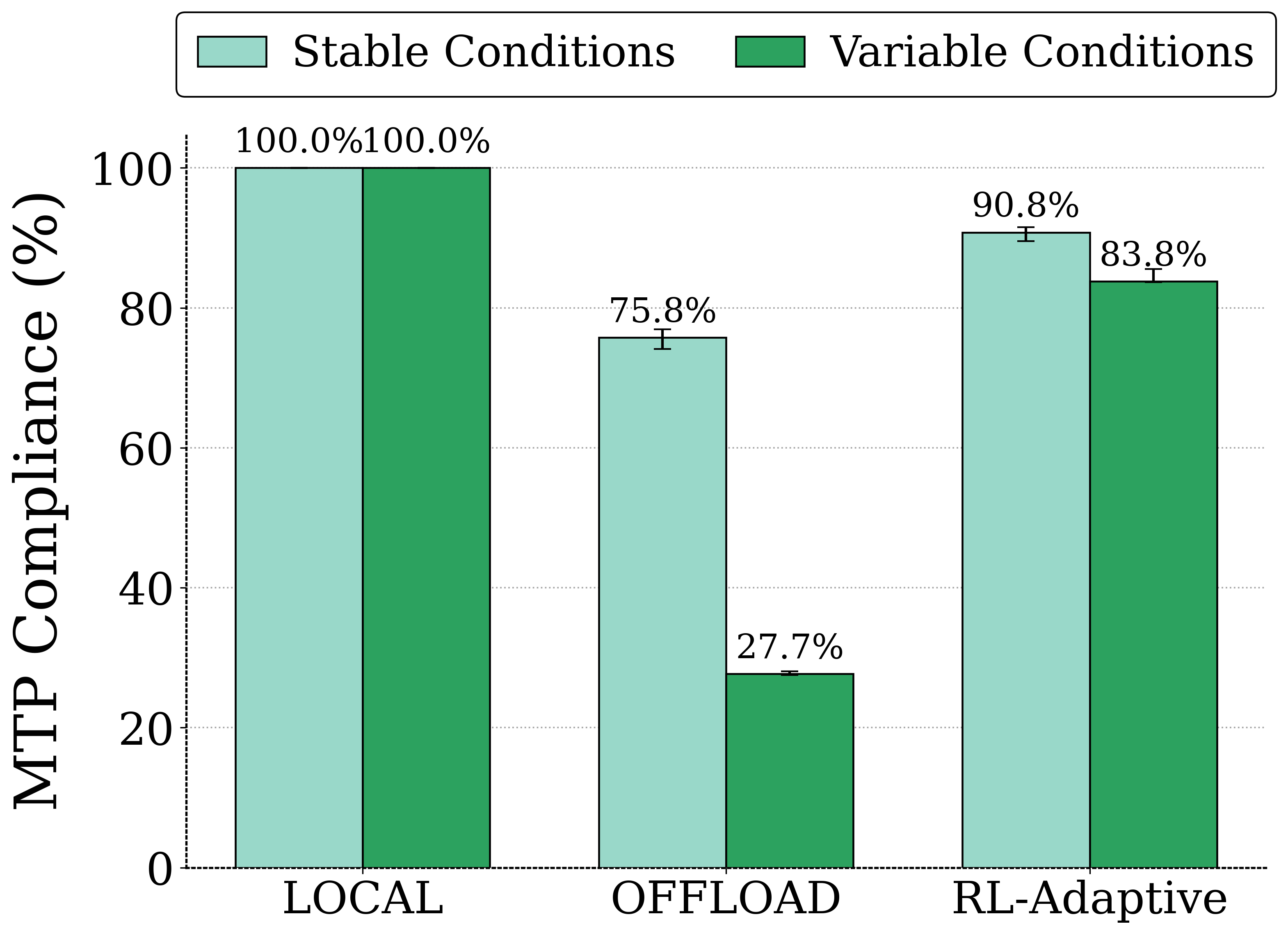}
        \caption{MTP compliance under stable and variable network conditions}
        \label{fig:result4b}
    \end{subfigure}

    \caption{MTP compliance of execution strategies under varying network conditions. The figure compares OFFLOAD and RL-Adaptive performance across bandwidth levels and shows overall strategy performance (LOCAL, OFFLOAD, RL-Adaptive) under stable and variable network scenarios.}
    \label{fig:result4}
    \vspace{-0.2in}
\end{figure}

Fig. \ref{fig:result3_compliance_bar} summarizes MTP latency compliance across bandwidth levels for OFFLOAD execution and the proposed RL-Adaptive strategy. OFFLOAD achieves only 60.8\% compliance at 1 Gbps and 66.5\% at 500 Mbps, with compliance dropping to near zero at 100 Mbps and below as transmission delays exceed the MTP budget. The relatively low compliance at high bandwidth occurs because bandwidth changes in cycles; when the network recovers from a low-bandwidth phase, delayed frames from the previous congested period temporarily raise latency even though bandwidth has increased. In contrast, the RL policy maintains 80–91\% compliance across all bandwidth levels, including 87.7\% at 1 Mbps, by dynamically switching between local and offloaded execution.

Fig. \ref{fig:result4b} compares MTP compliance under stable and variable network conditions. OFFLOAD fails to meet a satisfactory compliance level even under stable conditions (75.8\%) and degrades sharply under bandwidth variability, dropping to 27.7\%, a loss of 48.1 percentage points. In contrast, the RL-Adaptive strategy achieves 90.8\% compliance under stable conditions and maintains 83.8\% under variable bandwidth, degrading by only 7.0 percentage points. This represents a 6.9× smaller degradation compared to OFFLOAD, demonstrating that the learned policy effectively adapts to changing network conditions. The advantage of RL over OFFLOAD increases from 15.0\% under stable conditions to 56.1\% under variable bandwidth. While adaptive execution provides meaningful improvement even in stable environments, its benefit becomes decisive under network stress.

\begin{figure}[b]
\vspace{-0.2in}
    \centering
    \begin{subfigure}{0.492\columnwidth}
        \centering
        \includegraphics[width=\linewidth]{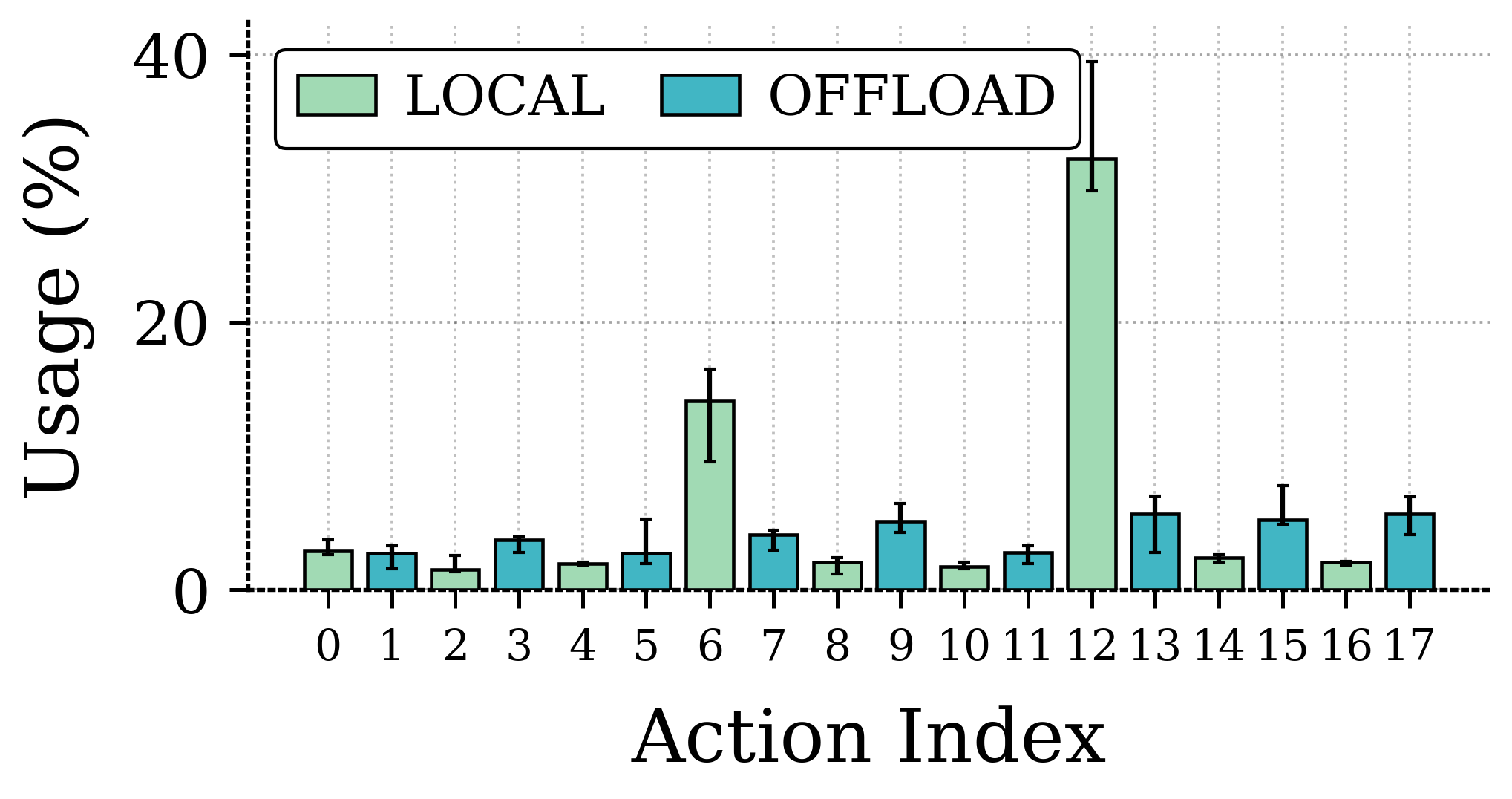}
        \caption{Stable network}
        \label{fig:result6a}
    \end{subfigure}
    \hfill
    \begin{subfigure}{0.492\columnwidth}
        \centering
        \includegraphics[width=\linewidth]{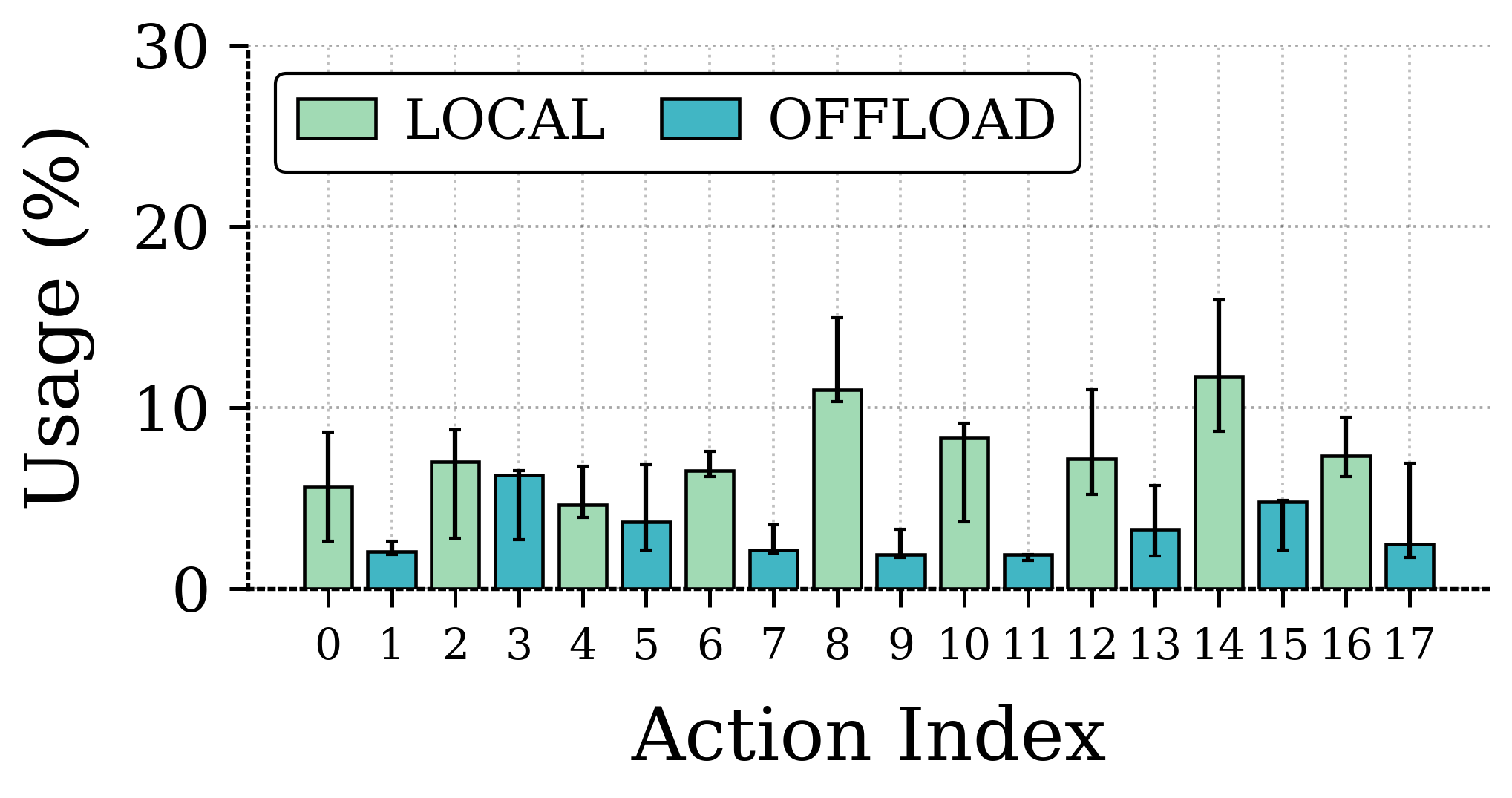}
        \caption{Disrupted Network}
        \label{fig:result6b}
    \end{subfigure}
    \caption{Learned policies for different network conditions. 
    }
    \label{fig:result6}
    \vspace{-0.1in}
\end{figure}

\subsubsection{Learned policy}
Fig. \ref{fig:result6} compares the action distributions learned by the RL controller under stable and variable network conditions. Each bar represents the median usage across three independent training seeds, with min–max error bars indicating policy variance. The action space encoding is shown in Tab. \ref{tab:action_space}.
Fig. \ref{fig:result6a} shows the action distribution learned under stable network conditions. Although distributed among the different policies, the highest used policy concentrates around Action 12 (low-frequency IMU, LOW quality, LOCAL) with Action 6 (medium-frequency IMU, LOW quality, LOCAL) as the next most common configuration. Offloading decisions appear across Actions 13, 15, and 17, which correspond to different image quality levels processed on the edge server. This pattern indicates a power-efficient strategy in which the agent minimizes local sensing cost while selectively offloading frames to maintain latency performance.

In contrast, Fig. \ref{fig:result6b} shows the policy under variable bandwidth conditions. The distribution becomes much more uniform and no single action dominates. The most common configurations include Actions 14 and 8 with MEDIUM image quality in LOCAL mode, along with higher-fidelity local settings such as Actions 16 and 10. The agent therefore relies more on local execution and increases sensing quality when offloading becomes unreliable. 

Across both conditions, the RL controller uses the full action space rather than collapsing to a single configuration. Under stable bandwidth the policy concentrates on a small set of power-efficient actions, whereas under variable conditions the distribution becomes more diverse.This behavior demonstrates that \emph{the learned policy internalizes the interaction between network variability, quality settings, and execution placement, a capability that is essential for sustained operation in closed-loop XR pipelines.}

\subsubsection{Peformance Comparison with Adaptive Baseline Strategies} 
To assess whether simpler adaptive controllers can match the proposed DQN policy, we compare RL-Adaptive against two baselines with different design philosophies. The GREEDY baseline selects the action with the highest estimated immediate reward at each decision step using the same reward function as RL, but without temporal planning, exploration, or replay. The THRESHOLD baseline applies a fixed rule, offloading when observed bandwidth exceeds 15 Mbps and otherwise executing locally, while always using maximum IMU rate and image quality. All three approaches use the same 1s decision interval, and results are reported as medians over three independent seeds with min–max envelopes or error bars.

\begin{figure}[t]
    \centering

    \begin{subfigure}{0.48\linewidth}
        \centering
        \includegraphics[width=\linewidth]{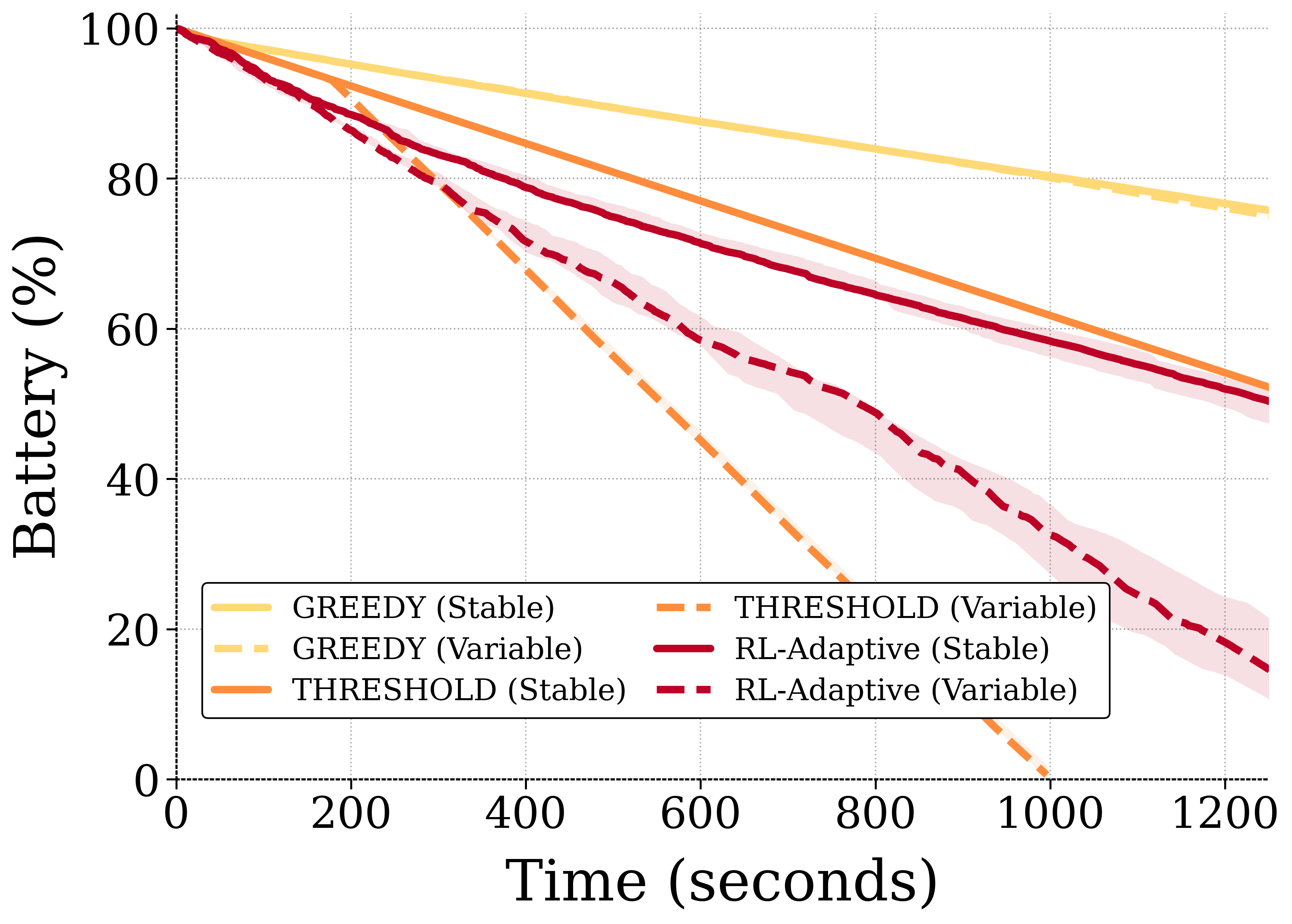}
        \caption{Battery depletion with time for RL-Adaptive and other adaptive approaches.}
        \label{fig:adaptive_baseline_battery_drain}
    \end{subfigure}
    \hspace{0.005in}
    \begin{subfigure}{0.48\linewidth}
        \centering
        \includegraphics[width=\linewidth]{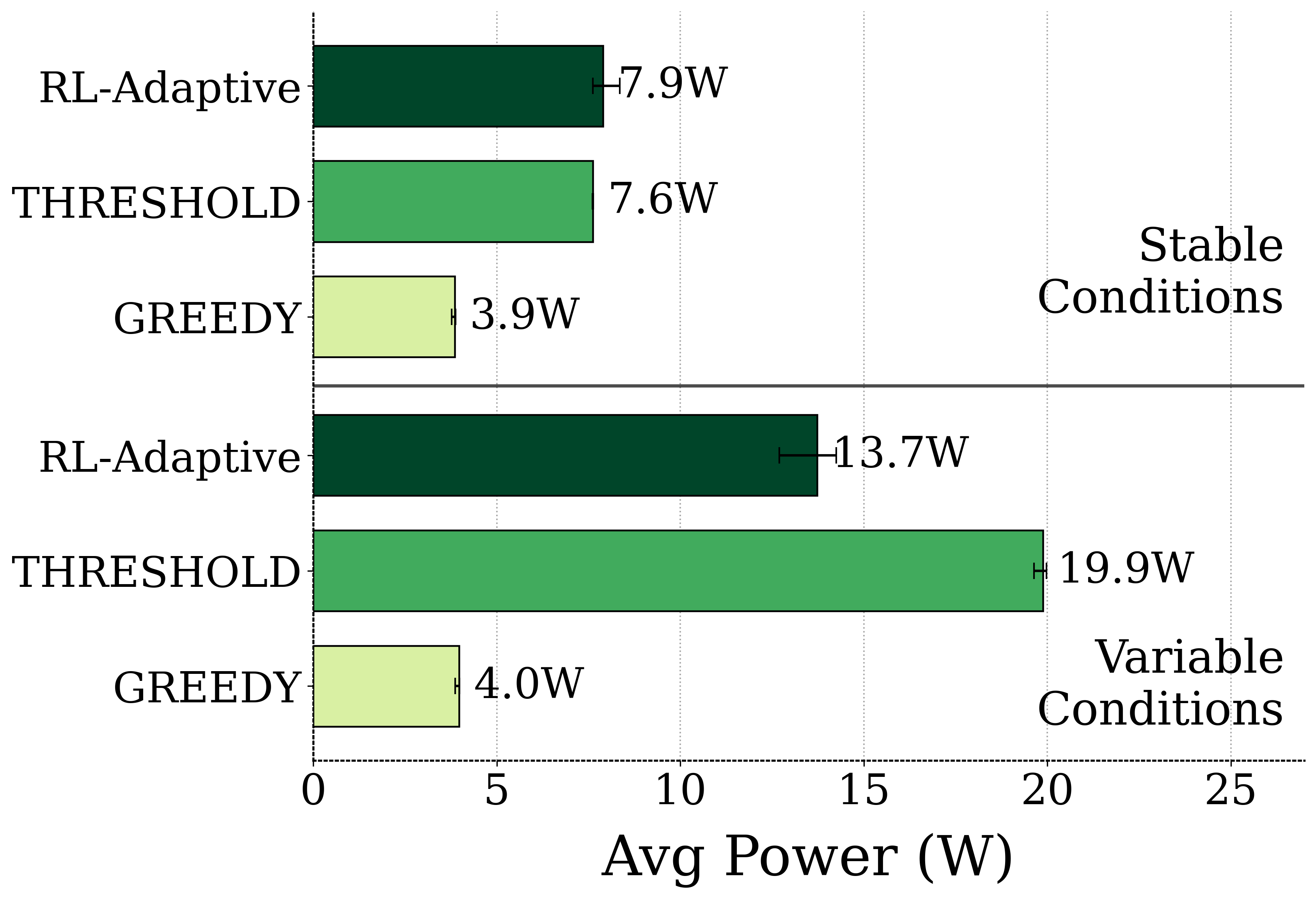}
        \caption{Average Power Consumption for RL-Adaptive and other adaptive approaches.}
        \label{fig:adaptive_baseline_power}
    \end{subfigure}
        \caption{Comparison of battery depletion and average power consumption among RL-Adaptive and other adaptive approaches over stable and variable network conditions.}
        \label{fig:adaptive_baseline_battery_power}
    \vspace{-0.2in}
\end{figure}

Figs. \ref{fig:adaptive_baseline_battery_drain} and \ref{fig:adaptive_baseline_power} compare both the battery drain trajectories and average power consumption across the three adaptive approaches under stable as well as variable bandwidth conditions. Under stable bandwidth, GREEDY exhibits the slowest battery drain, consuming about 3.9W, which results in a nearly flat battery curve over the experiment duration. THRESHOLD and RL-Adaptive consume 7.6W and 7.9W respectively, producing a moderate but steady battery decline while still completing the full evaluation window. Under variable bandwidth, the behavior diverges significantly. GREEDY remains largely unchanged at approximately 4.0W, leading to a battery trajectory similar to the stable case. In contrast, RL-Adaptive increases to 13.7W, resulting in a faster but still sustainable battery decline that allows the experiment to complete. THRESHOLD, however, experiences a dramatic rise in power consumption to 19.9 W, which accelerates battery drain and leads to complete battery depletion before the experiment finishes.

\begin{figure}[t]
    \centering

    \begin{subfigure}{0.5\linewidth}
        \centering
        \includegraphics[width=\linewidth]{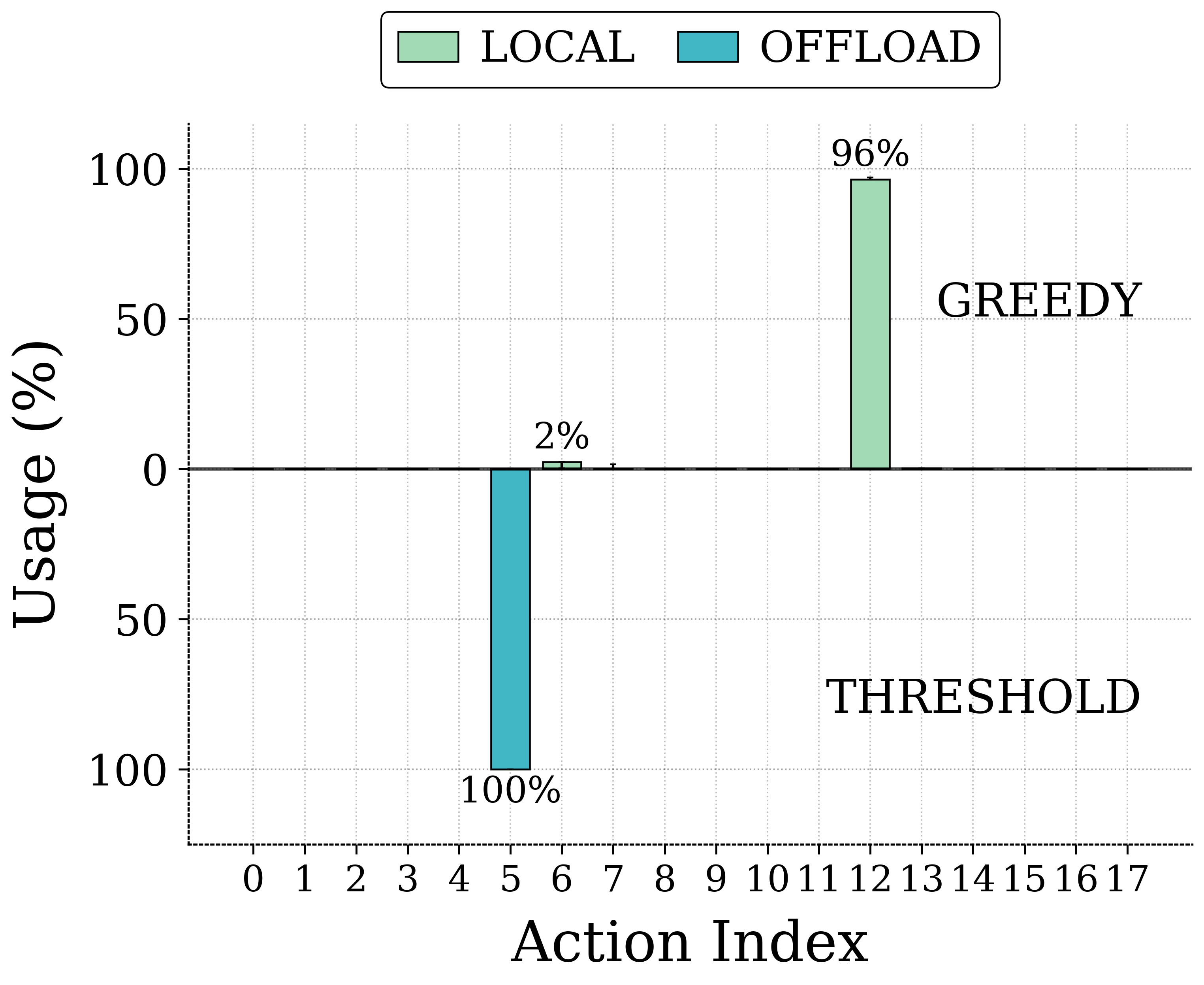}
        \caption{Used policies for GREEDY and THRESHOLD over stable network conditions.}
        \label{fig:adaptive_baseline_policy_stable}
    \end{subfigure}
    \hspace{-0.1in}
    \begin{subfigure}{0.5\linewidth}
        \centering
        \includegraphics[width=\linewidth]{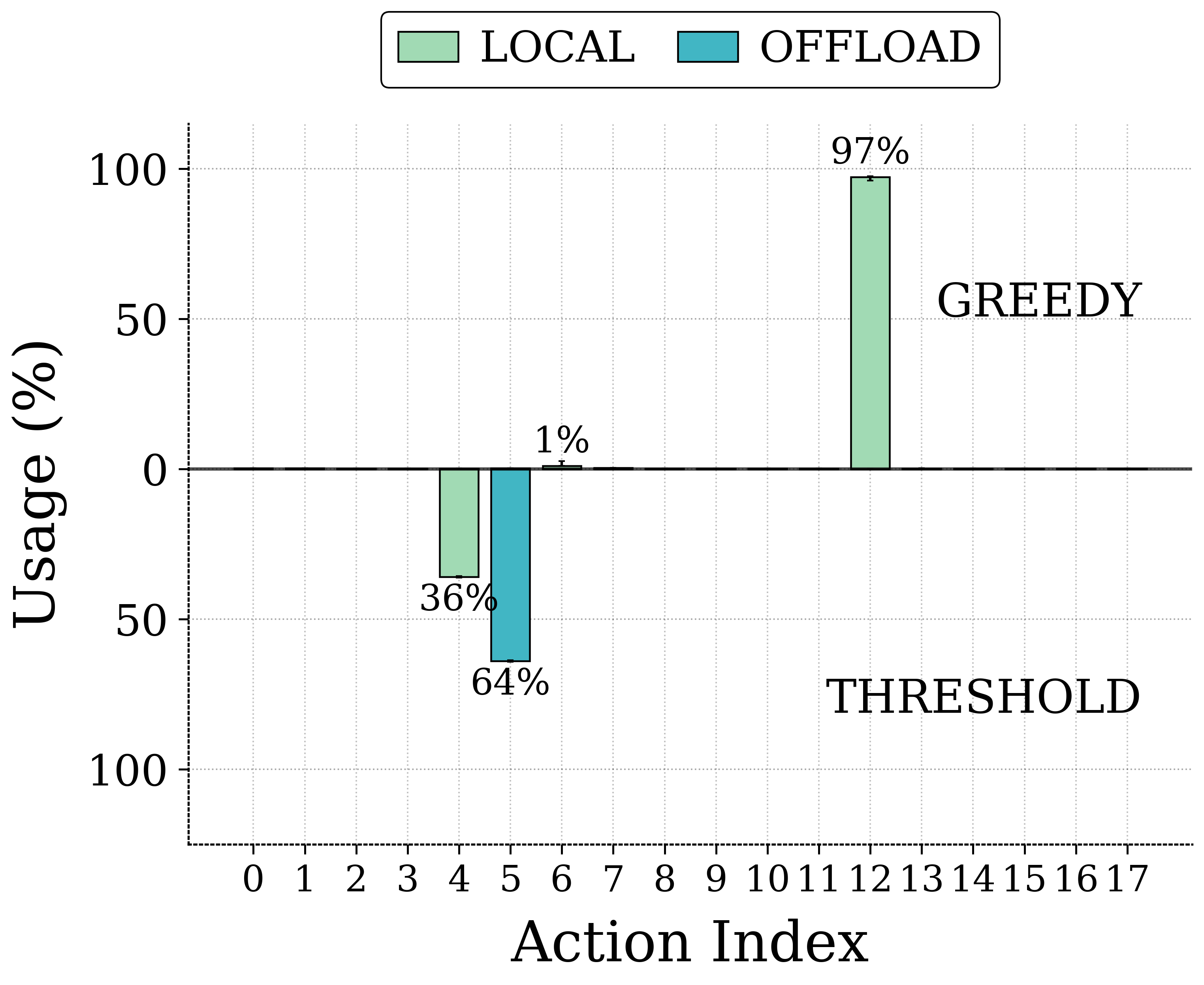}
        \caption{Used policies for GREEDY and THRESHOLD over variable network conditions.}
        \label{fig:adaptive_baseline_policy_variable}
    \end{subfigure}
        \caption{Comparison of policies used for GREEDY (top) and THRESHOLD (bottom) over stable (left) and variable (right) network conditions.}
    \vspace{-0.3in}
\end{figure}

The underlying cause of these results becomes clear when examining the policies learned by the different controllers over stable and variable network conditions, shown in Figs. \ref{fig:adaptive_baseline_policy_stable} and \ref{fig:adaptive_baseline_policy_variable}. GREEDY rapidly converges to a near-degenerate strategy dominated by a single configuration that uses LOW IMU rate and LOW image quality with LOCAL execution for almost all decisions. This configuration minimizes computational load, explaining the very low power consumption and slow battery drain observed in Fig. \ref{fig:adaptive_baseline_battery_power}. However, such aggressive reduction of sensing fidelity can degrade visual--inertial tracking reliability. VIO relies on sufficiently frequent inertial propagation and strong visual feature constraints for stable pose estimation \cite{RemoteVIO2025, liu2024slamhive}. More broadly, constantly reducing sensing fidelity to meet computational constraints necessarily sacrifices accuracy \cite{delmerico2018benchmark}. Since the VIO operates as a temporally coupled estimator, repeatedly operating at reduced sensing fidelity without corrective updates can accumulate drift over time \cite{sourya-vio}. This is precisely the limitation of GREEDY, which remains concentrated on the same low-resource configuration and rarely utilizes the edge server. 

THRESHOLD, in contrast, maintains HIGH IMU rate and HIGH image quality but adapts only the execution mode. When bandwidth drops and execution switches to LOCAL, this fixed high-fidelity configuration substantially increases computational demand, which explains the sharp rise in power consumption and the battery depletion observed under variable conditions. RL-Adaptive avoids both of these limitations by distributing decisions across the action space, dynamically selecting among LOW, MEDIUM, and HIGH sensing configurations while also adapting execution mode. As illustrated earlier in Fig. \ref{fig:result6}, \emph{RL-Adaptive interleaves lower-resource local actions with higher-fidelity local or offloaded actions depending on network conditions. Any drifts in the temporal filter state of VIO due to low quality sensing can be re-anchored with the help of higher-quality updates and drift accumulation can thus be limited, allowing RL-Adaptive to manage computational load without collapsing to a permanently degraded sensing configuration.}

\begin{figure*}[htbp]
    \centering
    \begin{subfigure}{0.155\textwidth}
        \centering
        \includegraphics[width=\linewidth]{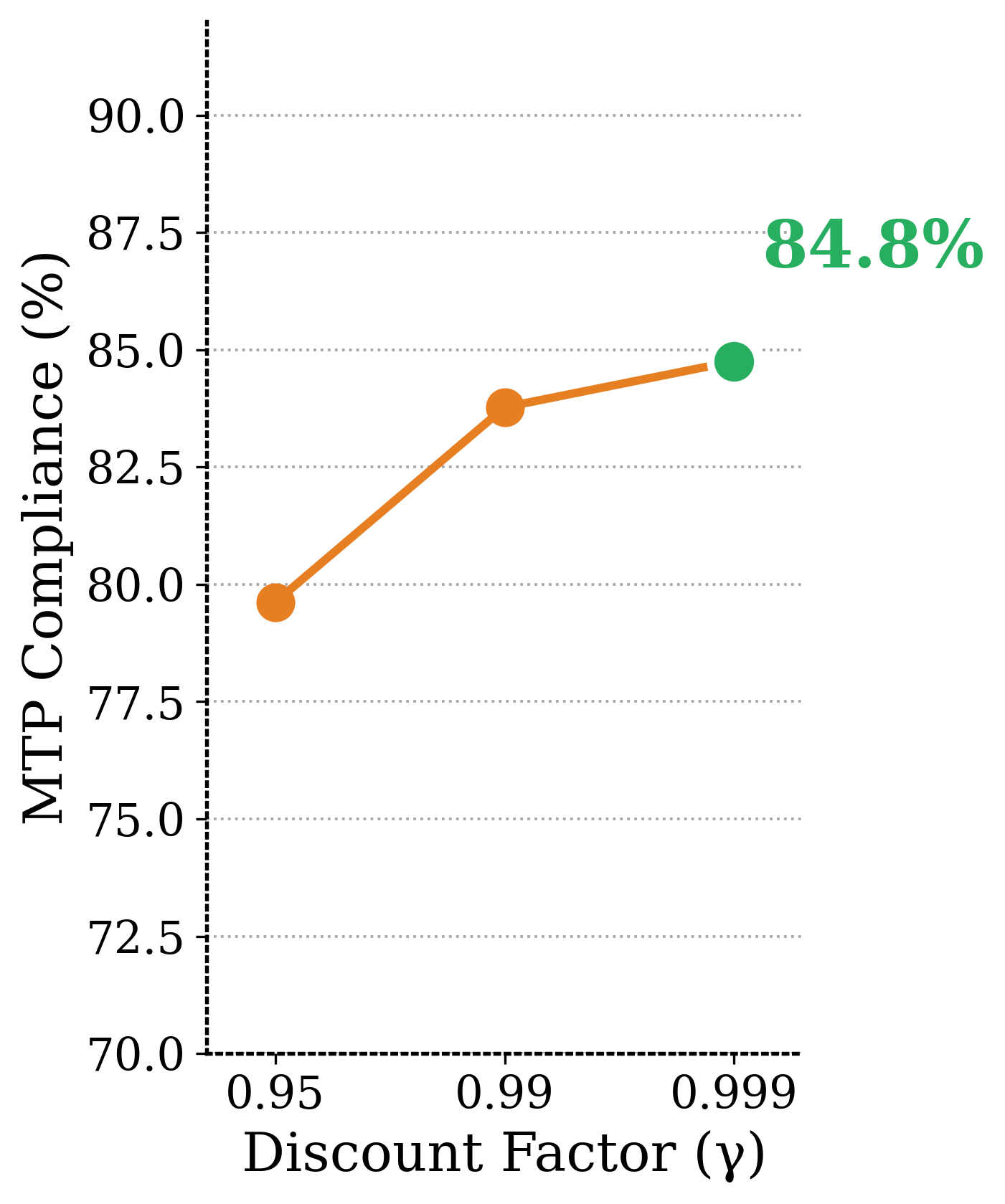}
        \caption{}
        \label{fig:result5a}
    \end{subfigure}
    \hfill
    \begin{subfigure}{0.155\textwidth}
        \centering
        \includegraphics[width=\linewidth]{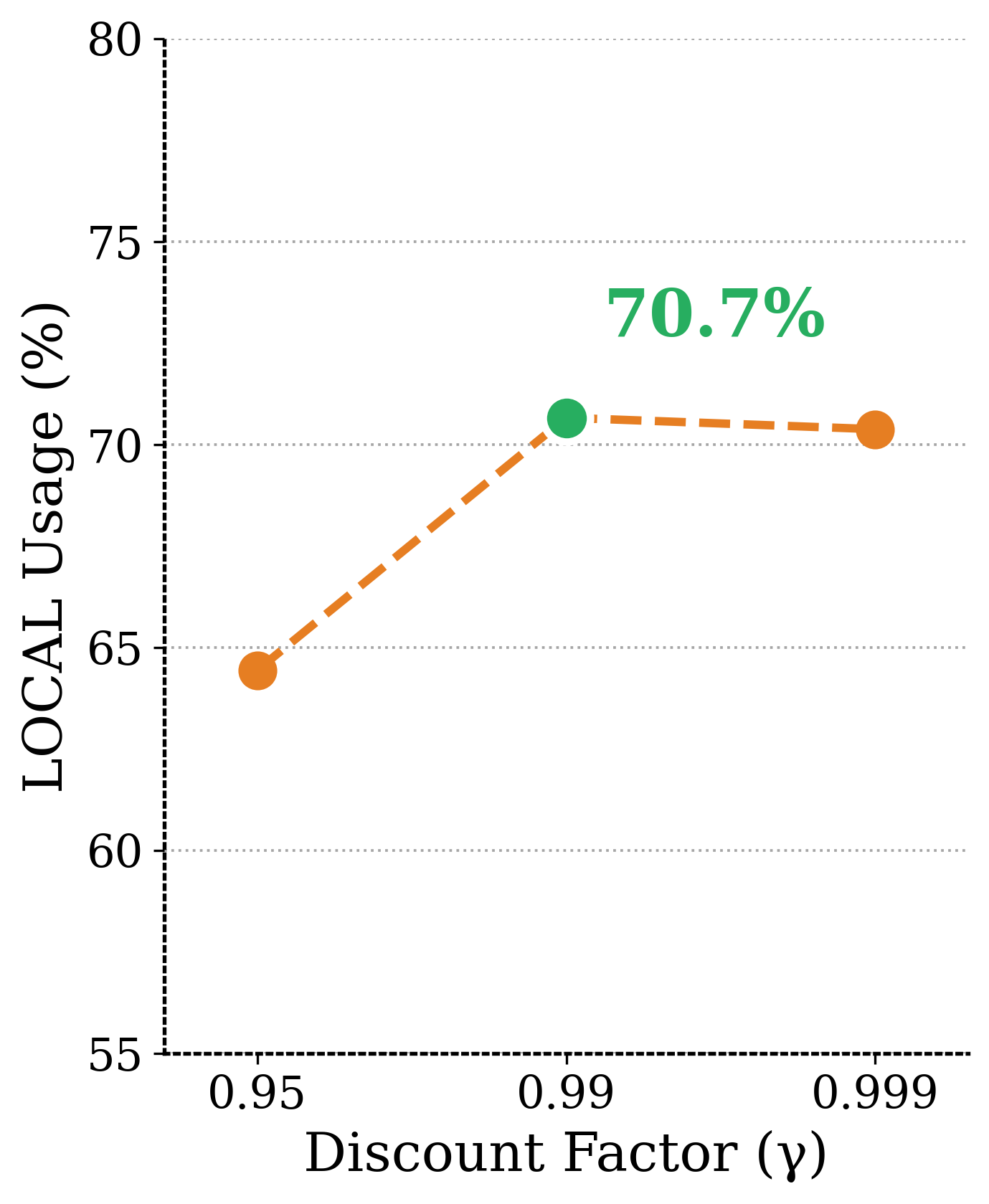}
        \caption{}
        \label{fig:result5d}
    \end{subfigure}
    \hfill
    \begin{subfigure}{0.155\textwidth}
        \centering
        \includegraphics[width=\linewidth]{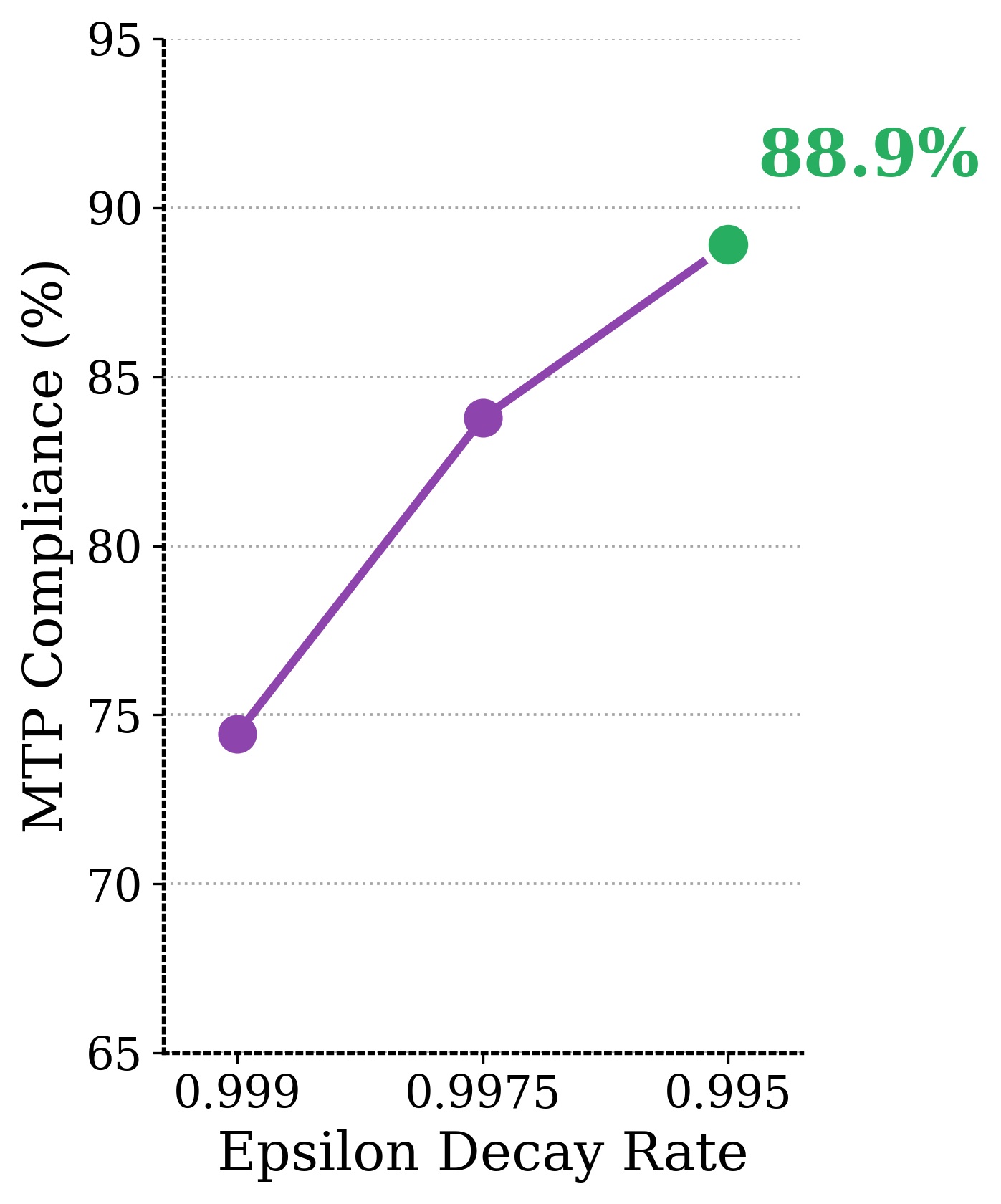}
        \caption{}
        \label{fig:result5b}
    \end{subfigure}
    \hfill
    \begin{subfigure}{0.155\textwidth}
        \centering
        \includegraphics[width=\linewidth]{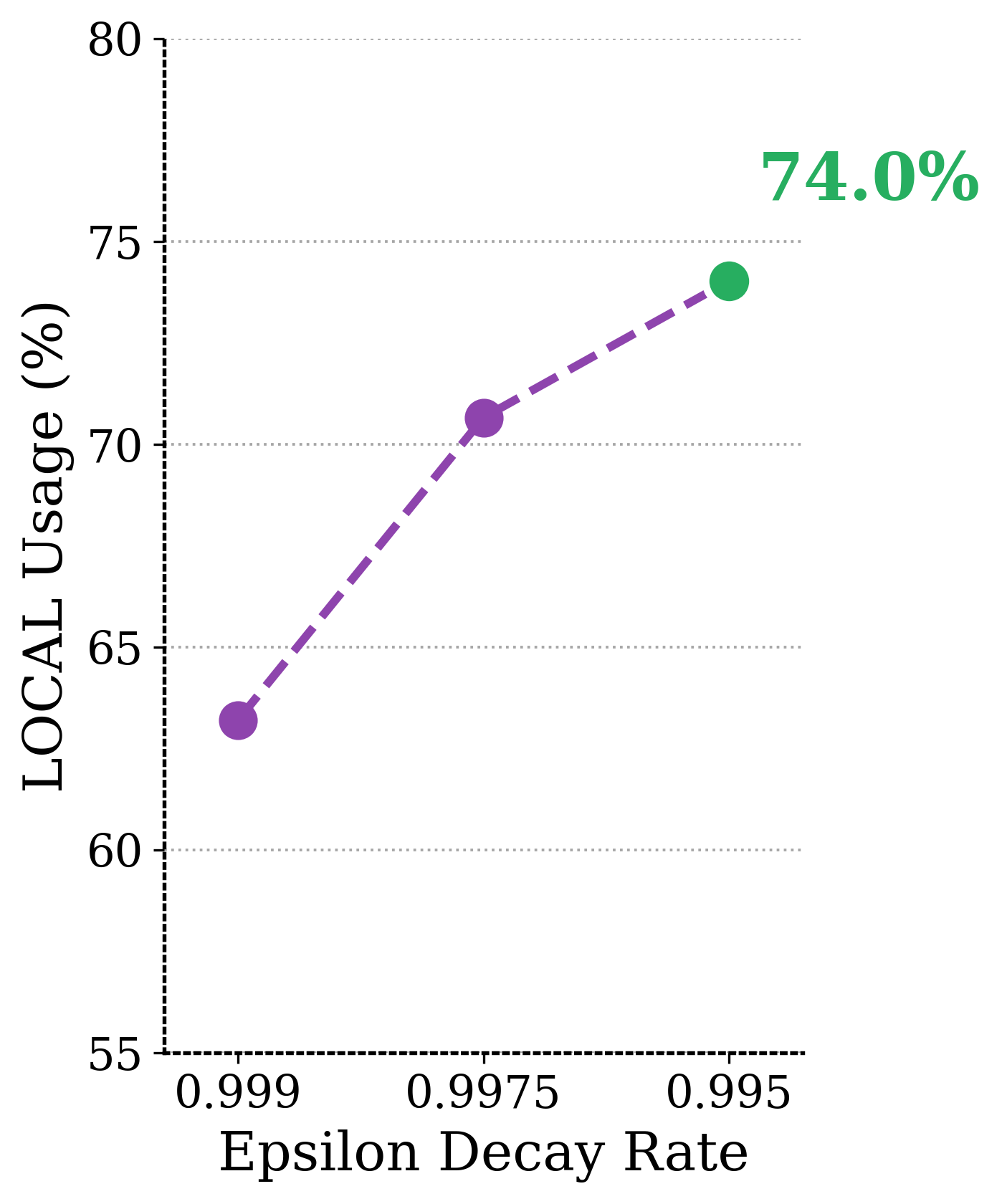}
        \caption{}
        \label{fig:result5e}
    \end{subfigure}
    \hfill
    \begin{subfigure}{0.155\textwidth}
        \centering
        \includegraphics[width=\linewidth]{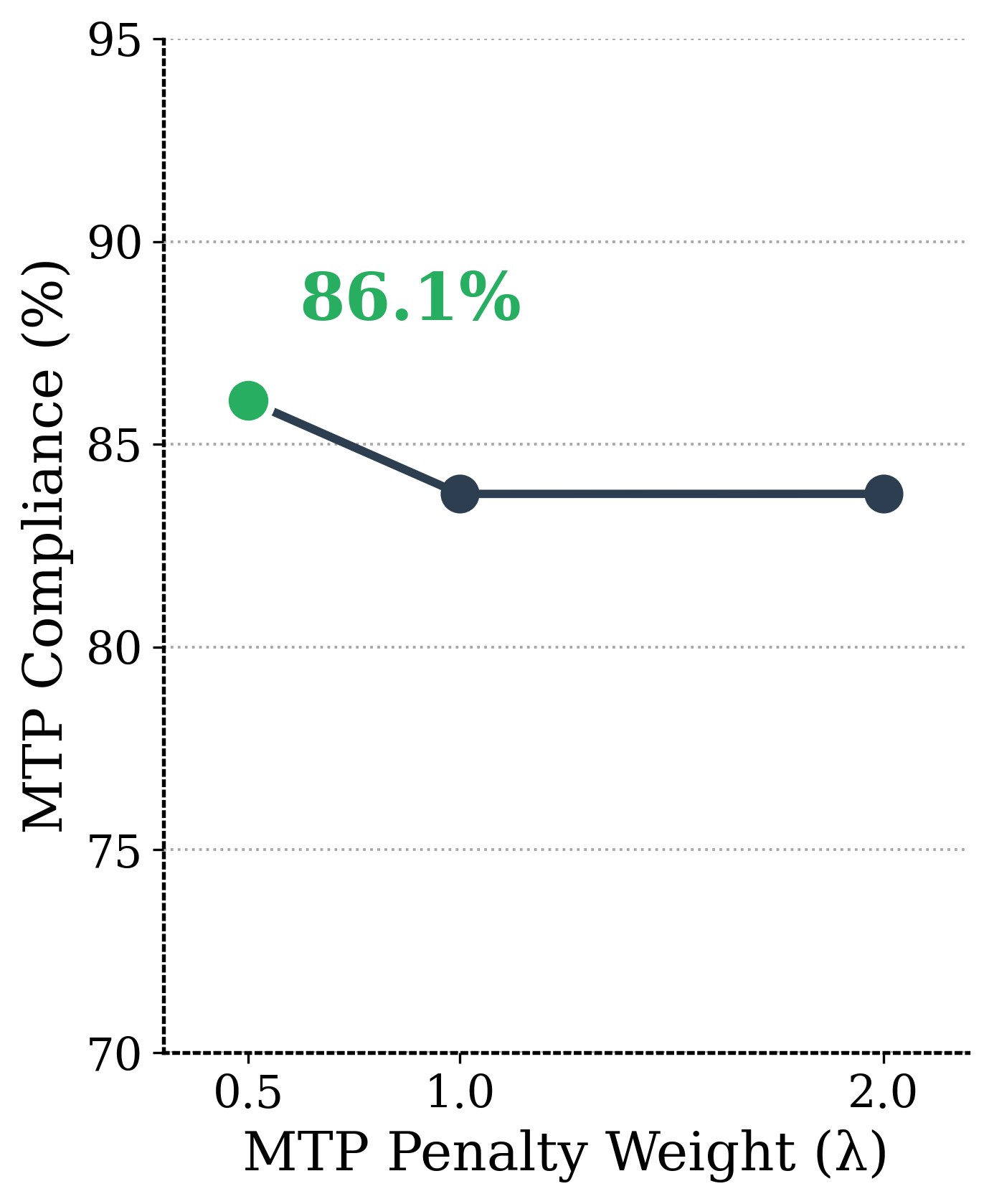}
        \caption{}
        \label{fig:result5c}
    \end{subfigure}
    \hfill
    \begin{subfigure}{0.155\textwidth}
        \centering
        \includegraphics[width=\linewidth]{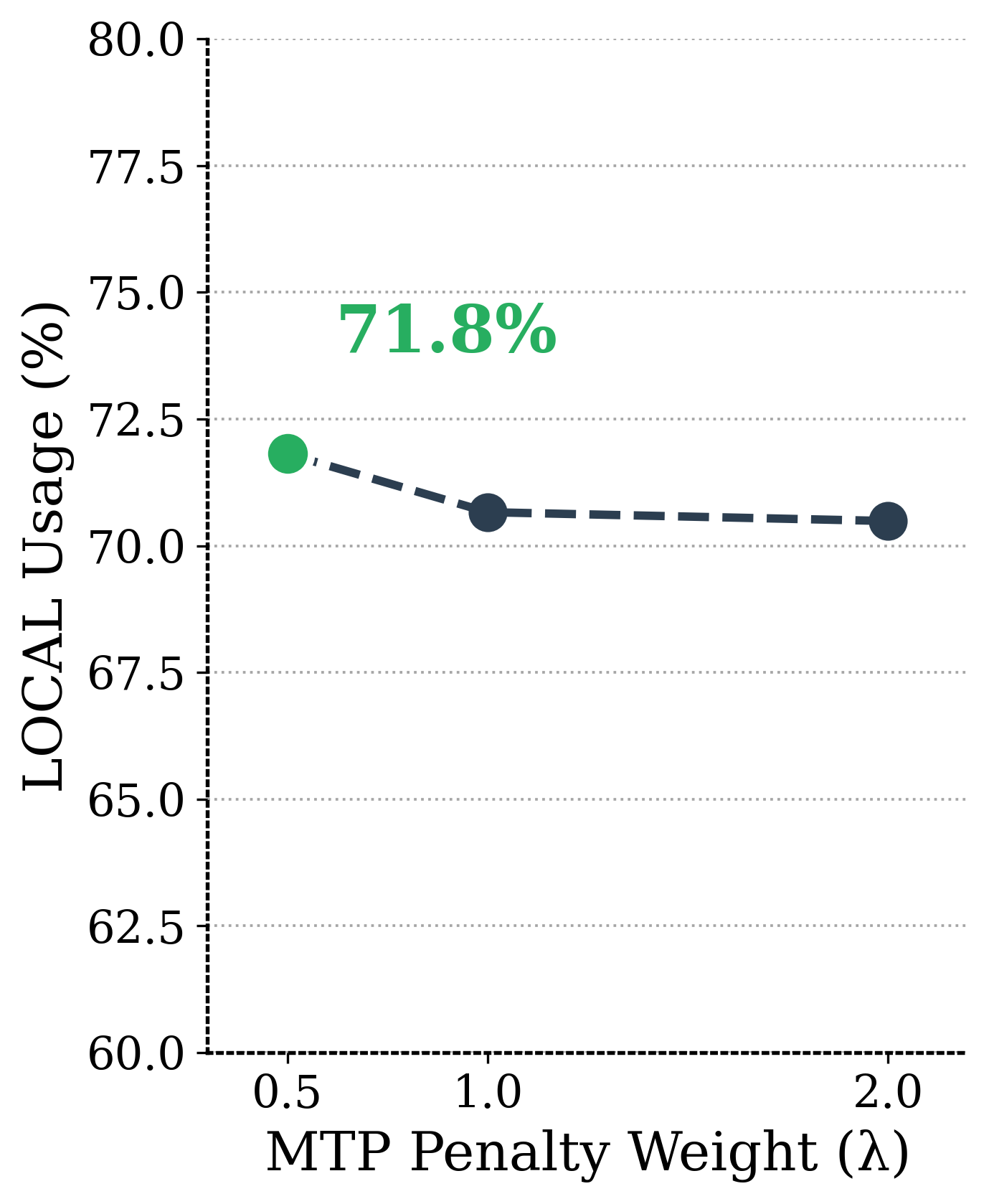}
        \caption{}
        \label{fig:result5f}
    \end{subfigure}
    \caption{Hyperparameter sensitivity of the DQN policy under variable bandwidth conditions. (a) Effect of $\gamma$, (b) Percentage Local Processing under different $\gamma$, (c) Effect of epsilon decay rate, (d) Percentage Local Processing under different epsilon, (e) Effect of MTP penalty weight, (f) Percentage Local Processing under different MTP penalty weights.}
    \label{fig:result5}
    \vspace{-0.2in}
\end{figure*}

\subsubsection{Hyperparameter Sensitivity Analysis}
\emph{Firstly}, we analyze the sensitivity of the RL controller to three key hyperparameters including the MTP penalty weight $\lambda$, the epsilon ($\epsilon$) decay rate, and the discount factor $\gamma$. For each parameter, the value is varied while the remaining parameters remain fixed at the baseline configuration $\lambda=1.0$, $\epsilon$-decay$=0.9975$, and $\gamma=0.99$.
The discount factor shows moderate sensitivity as illustrated in Figs. \ref{fig:result5a} and \ref{fig:result5d}. A short-sighted configuration with $\gamma=0.95$ achieves 79.6\% compliance with 64.4\% LOCAL execution. Increasing $\gamma$ to the baseline value 0.99 improves performance to 83.8\% compliance with 70.7\% LOCAL usage. Further increasing the value to $\gamma=0.999$ produces only a small improvement to 84.8\% compliance with 70.4\% LOCAL execution, indicating diminishing returns beyond $\gamma=0.99$.

The exploration schedule ($\epsilon$-decay) has the largest influence on performance as shown in Figs. \ref{fig:result5b} and \ref{fig:result5e}. Slow decay (0.999) results in 74.4\% compliance with 63.2\% LOCAL execution, indicating that the agent continues exploring for most of the session. The baseline $\epsilon$ of 0.9975 achieves 83.8\% compliance with 70.7\% LOCAL usage. Faster decay (0.995) improves performance further to 88.9\% compliance with 74.0\% LOCAL execution, producing a 14.5 percentage point compliance range across configurations. These results indicate that faster convergence to exploitation significantly improves cumulative performance when learning occurs within a single session.

The MTP penalty weight is the least sensitive parameter as shown in Figs. \ref{fig:result5c} and \ref{fig:result5f}. A weaker penalty $\gamma=0.5$ produces 86.1\% compliance with 71.8\% LOCAL execution. The baseline configuration $\gamma=1.0$ achieves 83.8\% compliance with 70.7\% LOCAL usage, while increasing the penalty to $\gamma=2.0$ produces nearly identical performance with 83.8\% compliance and 70.5\% LOCAL execution. The resulting 2.3 percentage point compliance range indicates that performance is largely unchanged across these penalty settings.

\begin{table}[t]
\centering
\caption{Hyperparameter Sensitivity Analysis under variable network conditions}
\label{tab:decision_frequency}

\setlength{\tabcolsep}{3pt}
\renewcommand{\arraystretch}{1.2}

\begin{tabular}{|l|c|c|c|}
\hline
\scriptsize\textbf{\makecell{Decision \\ Interval}} &
\scriptsize\textbf{\makecell{MTP \\ Compliance}} &
\scriptsize\textbf{\makecell{Avg \\ Power (W)}} &
\scriptsize\textbf{\makecell{LOCAL \\ (\%)}} \\
\hline
\scriptsize 1.0 s (baseline) & \scriptsize 83.8\% & \scriptsize 13.7 & \scriptsize 70.0 \\
\hline
\scriptsize 0.75 s           & \scriptsize 87.8\% & \scriptsize 15.5 & \scriptsize 72.9 \\
\hline
\scriptsize 0.5 s            & \scriptsize 84.0\% & \scriptsize 15.3 & \scriptsize 66.2 \\
\hline
\end{tabular}
\vspace{-0.2in}
\end{table}

\emph{Secondly}, we evaluate the sensitivity of RL-Adaptive to the decision interval by reducing the baseline frequency from 1.0 s to 0.75 s and 0.5 s under variable bandwidth conditions. As summarized in Table \ref{tab:decision_frequency}, the 0.75 s interval provides the best performance, as the controller can react more quickly to changing conditions while still allowing sufficient time for the VIO pipeline to reflect the impact of previous actions. In contrast, the 0.5 s interval does not improve performance because decisions begin to occur faster than the system’s feedback latency, causing unstable policy updates and oscillatory behavior.

\begin{table}[t]
\centering
\caption{Action Space Encoding}
\label{tab:action_space}

\setlength{\tabcolsep}{3pt}
\renewcommand{\arraystretch}{1.2}

\begin{tabular}{|c|c|c|c|}
\hline
\scriptsize\textbf{Action} & \scriptsize\textbf{IMU Rate} & \scriptsize\textbf{Quality} & \scriptsize\textbf{Mode} \\
\hline
\scriptsize 0  & \scriptsize HIGH   & \scriptsize LOW    & \scriptsize LOCAL \\
\hline
\scriptsize 1  & \scriptsize HIGH   & \scriptsize LOW    & \scriptsize OFFLOAD \\
\hline
\scriptsize 2  & \scriptsize HIGH   & \scriptsize MEDIUM & \scriptsize LOCAL \\
\hline
\scriptsize 3  & \scriptsize HIGH   & \scriptsize MEDIUM & \scriptsize OFFLOAD \\
\hline
\scriptsize 4  & \scriptsize HIGH   & \scriptsize HIGH   & \scriptsize LOCAL \\
\hline
\scriptsize 5  & \scriptsize HIGH   & \scriptsize HIGH   & \scriptsize OFFLOAD \\
\hline
\scriptsize 6  & \scriptsize MEDIUM & \scriptsize LOW    & \scriptsize LOCAL \\
\hline
\scriptsize 7  & \scriptsize MEDIUM & \scriptsize LOW    & \scriptsize OFFLOAD \\
\hline
\scriptsize 8  & \scriptsize MEDIUM & \scriptsize MEDIUM & \scriptsize LOCAL \\
\hline
\scriptsize 9  & \scriptsize MEDIUM & \scriptsize MEDIUM & \scriptsize OFFLOAD \\
\hline
\scriptsize 10 & \scriptsize MEDIUM & \scriptsize HIGH   & \scriptsize LOCAL \\
\hline
\scriptsize 11 & \scriptsize MEDIUM & \scriptsize HIGH   & \scriptsize OFFLOAD \\
\hline
\scriptsize 12 & \scriptsize LOW    & \scriptsize LOW    & \scriptsize LOCAL \\
\hline
\scriptsize 13 & \scriptsize LOW    & \scriptsize LOW    & \scriptsize OFFLOAD \\
\hline
\scriptsize 14 & \scriptsize LOW    & \scriptsize MEDIUM & \scriptsize LOCAL \\
\hline
\scriptsize 15 & \scriptsize LOW    & \scriptsize MEDIUM & \scriptsize OFFLOAD \\
\hline
\scriptsize 16 & \scriptsize LOW    & \scriptsize HIGH   & \scriptsize LOCAL \\
\hline
\scriptsize 17 & \scriptsize LOW    & \scriptsize HIGH   & \scriptsize OFFLOAD \\
\hline
\end{tabular}

\vspace{-0.2in}
\end{table}

\subsubsection{Overhead of the RL controller}
To justify describing the controller as \emph{lightweight}, we measure its runtime overhead by comparing system power under two configurations that are identical in all aspects except the RL controller plugin. As documented in Tab. \ref{tab:dqn_overhead}, in both configurations, the system runs on LOCAL with HIGH IMU and HIGH camera quality while loading the same infrastructure plugins. In the RL configuration, the RL-Adaptive controller plugin is enabled, but its selected action is overridden so that execution mode and sensor settings remain identical, ensuring that the only difference between the two systems is the presence of the DQN performing inference and online training. Each configuration is executed with three independent seeds, and the mean results are reported in Tab.~\ref{tab:dqn_overhead}. Enabling the controller increases average system power from 22.4W to 23.3W, corresponding to a 0.9W overhead (4.0\%). Importantly, this overhead is small relative to the energy savings achieved by the RL policy itself, which reduces client power consumption by several watts through adaptive offloading and sensor configuration decisions. Decision latency remains small relative to the 1s control interval. The 157$\mu$s median latency corresponds to inference-only steps, while the 3.945ms P95 latency corresponds to decisions that include replay-buffer training and gradient updates. Even at P95 the controller occupies only 0.4\% of the decision interval, confirming that the RL-Adaptive controller introduces minimal runtime overhead. All evaluation results and related codes are available through Github~\cite{git}.

\begin{table}[t]
\centering
\caption{Measured overhead and architecture of the RL-Adaptive controller}
\label{tab:dqn_overhead}

\setlength{\tabcolsep}{3pt}
\renewcommand{\arraystretch}{1.2}

\begin{tabular}{|p{3.1cm}|>{\centering\arraybackslash}p{3.2cm}|}
\hline
\scriptsize\textbf{Metric} & \scriptsize\textbf{Value} \\
\hline

\scriptsize LOCAL baseline & \scriptsize 20.7 $\pm$ 0.1 W \\
\hline
\scriptsize LOCAL + plugins & \scriptsize 22.4 $\pm$ 0.1 W \\
\hline
\scriptsize \makecell[l]{LOCAL + plugins \\ + RL-Adaptive} & \scriptsize 23.3 $\pm$ 0.1 W \\
\hline
\scriptsize RL controller overhead & \scriptsize 0.9 W (4.0\%) \\
\hline

\multicolumn{2}{|l|}{\scriptsize\textbf{Decision latency}} \\
\hline
\scriptsize Median inference latency & \scriptsize 157 $\mu$s \\
\hline
\scriptsize Mean decision latency & \scriptsize 1247 $\mu$s \\
\hline
\scriptsize P95 latency (training step) & \scriptsize 3945 $\mu$s \\
\hline
\scriptsize Decision interval & \scriptsize 1 s \\
\hline
\scriptsize P95 interval utilization & \scriptsize 0.4\% \\
\hline

\multicolumn{2}{|l|}{\scriptsize\textbf{DQN architecture}} \\
\hline
\scriptsize Network structure & \scriptsize MLP (5–128–128–18) \\
\hline
\scriptsize Hidden activation & \scriptsize ReLU \\
\hline
\scriptsize Parameters (Q-network) & \scriptsize 19,602 \\
\hline
\scriptsize Target network & \scriptsize identical copy \\
\hline
\scriptsize Decision frequency & \scriptsize 1 Hz \\
\hline
\end{tabular}
\vspace{-0.2in}
\end{table}

\section{Conclusions and Future Work}
\label{sec:conclusion}

This paper examined the challenge of managing execution placement for immersive XR workloads that must maintain high MTP latency compliance while operating under limited battery capacity on end-devices. Our analysis shows that fixed execution strategies are fundamentally limited, either trading responsiveness for battery life or becoming highly sensitive to network variability, while simple adaptive heuristics also fail to balance these competing objectives. To address this trade-off, we introduced a battery-aware execution management framework that jointly adapts execution placement, workload configuration, and sensing quality while accounting for latency requirements and battery dynamics. The framework employs a lightweight RL controller that operates online without requiring offline training and adds minimal runtime overhead. Through evaluation under representative XR workloads and variable network conditions, we showed that the RL-Adaptive policy extended projected battery life by approximately 163\% compared to static local execution while maintaining over 90\% MTP compliance under 
stable conditions and 83.8\% under variable bandwidth. These results highlight the importance of explicitly managing latency–energy interactions in closed-loop XR pipelines. At the same time, the current design provides empirical latency compliance rather than formal guarantees and considers a single client–edge deployment; extending the framework to multi-device environments with shared edge resources and exploring constrained RL methods to provide stronger compliance guarantees are some important directions for future work.

\bibliographystyle{ieeetr}
\bibliography{refs}

@article{bassbouss2016high,
  title={High quality 360 video rendering and streaming},
  author={Bassbouss, Louay and Steglich, Stephan and Lasak, Martin},
  journal={NEM Summit proceedings},
  year={2016}
}

@article{Dong2024TaskOffloadingSurvey,
title = {Task offloading strategies for mobile edge computing: A survey},
journal = {Computer Networks},
volume = {254},
pages = {110791},
year = {2024},
issn = {1389-1286},
doi = {https://doi.org/10.1016/j.comnet.2024.110791},
url = {https://www.sciencedirect.com/science/article/pii/S1389128624006236},
author = {Shi Dong and Junxiao Tang and Khushnood Abbas and Ruizhe Hou and Joarder Kamruzzaman and Leszek Rutkowski and Rajkumar Buyya}
}

@article{Alwabel2025LatencyAware,
  title={Latency-Aware Task Offloading Mechanism for Mobile Edge Computing},
  author={Alwabel, Abdulelah},
  journal={CLOUD COMPUTING 2025},
  pages={105},
  year={2025}
}

@inproceedings{Gao2025XRgo,
author = {Gao, Steven and Liu, Jeffrey and Jiang, Qinjun and Sinclair, Finn and Sentosa, William and Godfrey, Brighten and Adve, Sarita},
title = {XRgo: Design and Evaluation of Rendering Offload for Low-Power Extended Reality Devices},
year = {2025},
isbn = {9798400714672},
publisher = {Association for Computing Machinery},
address = {New York, NY, USA},
url = {https://doi.org/10.1145/3712676.3714444},
doi = {10.1145/3712676.3714444},
booktitle = {Proceedings of the 16th ACM Multimedia Systems Conference},
pages = {124–135},
numpages = {12},
location = {Stellenbosch, South Africa},
series = {MMSys '25}
}

@inproceedings{Ometov2023EdgeEnergy,
  title={Will Edge Computing Enable Location based Extended/Mixed Reality Mobile Gaming? Demystifying Trade-off of Execution Time vs. Energy Consumption},
  author={Aleksandr Ometov and Jari Nurmi},
  booktitle={WIPHAL},
  year={2023},
  url={https://api.semanticscholar.org/CorpusID:260442348}
}

@misc{Duru2025ResourceAllocationXR,
      title={Resource Allocation for XR with Edge Offloading: A Reinforcement Learning Approach}, 
      author={Alperen Duru and Mohammad Mozaffari and Ticao Zhang and Mehrnaz Afshang},
      year={2025},
      eprint={2510.22505},
      archivePrefix={arXiv},
      primaryClass={cs.IT},
      url={https://arxiv.org/abs/2510.22505}, 
}

@INPROCEEDINGS{illixr,
  author={Huzaifa, Muhammad and Desai, Rishi and Grayson, Samuel and Jiang, Xutao and Jing, Ying and Lee, Jae and Lu, Fang and Pang, Yihan and Ravichandran, Joseph and Sinclair, Finn and Tian, Boyuan and Yuan, Hengzhi and Zhang, Jeffrey and Adve, Sarita V.},
  booktitle={2021 IEEE International Symposium on Workload Characterization (IISWC)}, 
  title={ILLIXR: Enabling End-to-End Extended Reality Research}, 
  year={2021},
  volume={},
  number={},
  pages={24-38},
  doi={10.1109/IISWC53511.2021.00014}}

@INPROCEEDINGS{ZhangDebroy2020,
  author={Zhang, Xiaojie and Debroy, Saptarshi},
  booktitle={ICC 2020 - 2020 IEEE International Conference on Communications (ICC)}, 
  title={Energy Efficient Task Offloading for Compute-intensive Mobile Edge Applications}, 
  year={2020},
  volume={},
  number={},
  pages={1-6},
  doi={10.1109/ICC40277.2020.9149012}}

@INPROCEEDINGS{ZhangEtAl2021,
  author={Zhang, Siqi and Yi, Na and Ma, Yi},
  booktitle={2021 IEEE 93rd Vehicular Technology Conference (VTC2021-Spring)}, 
  title={Correlation-Based Device Energy-Efficient Dynamic Multi-Task Offloading for Mobile Edge Computing}, 
  year={2021},
  volume={},
  number={},
  pages={1-5},
  doi={10.1109/VTC2021-Spring51267.2021.9448864}}

@ARTICLE{TangWong2020,
  author={Tang, Ming and Wong, Vincent W.S.},
  journal={IEEE Transactions on Mobile Computing}, 
  title={Deep Reinforcement Learning for Task Offloading in Mobile Edge Computing Systems}, 
  year={2022},
  volume={21},
  number={6},
  pages={1985-1997},
  doi={10.1109/TMC.2020.3036871}}

@inproceedings{RemoteVIO2025,
author = {Jiang, Qinjun and Pang, Yihan and Sentosa, William and Gao, Steven and Huzaifa, Muhammad and Zhang, Jeffrey and Perez-Ramirez, Javier and Das, Dibakar and Gonzalez-Aguirre, David and Godfrey, Brighten and Adve, Sarita},
title = {RemoteVIO: Offloading Head Tracking in an End-to-End XR System},
year = {2025},
isbn = {9798400714672},
publisher = {Association for Computing Machinery},
address = {New York, NY, USA},
url = {https://doi.org/10.1145/3712676.3714442},
doi = {10.1145/3712676.3714442},
booktitle = {Proceedings of the 16th ACM Multimedia Systems Conference},
pages = {101–112},
numpages = {12},
location = {Stellenbosch, South Africa},
series = {MMSys '25}
}

@misc{Dixit2023MTP,
      title={Minimizing the Motion-to-Photon-delay (MPD) in Virtual Reality Systems}, 
      author={Akanksha Dixit and Smruti R. Sarangi},
      year={2023},
      eprint={2301.10408},
      archivePrefix={arXiv},
      primaryClass={cs.AR},
      url={https://arxiv.org/abs/2301.10408}, 
}

@inproceedings{Zhu2022EdgeVR,
author = {Zhu, Ziehen and Feng, Xianglong and Tang, Zhongze and Jiang, Nan and Guo, Tian and Xu, Lisong and Wei, Sheng},
title = {Power-efficient live virtual reality streaming using edge offloading},
year = {2022},
isbn = {9781450393836},
publisher = {Association for Computing Machinery},
address = {New York, NY, USA},
url = {https://doi.org/10.1145/3534088.3534351},
doi = {10.1145/3534088.3534351},
pages = {57–63},
numpages = {7},
series = {NOSSDAV '22}
}

@article{Casasnovas2024EdgeCloudVR,
   title={Experimental evaluation of interactive Edge/Cloud Virtual Reality gaming over Wi-Fi using unity render streaming},
   volume={226–227},
   ISSN={0140-3664},
   url={http://dx.doi.org/10.1016/j.comcom.2024.08.001},
   DOI={10.1016/j.comcom.2024.08.001},
   journal={Computer Communications},
   publisher={Elsevier BV},
   author={Casasnovas, Miguel and Michaelides, Costas and Carrascosa-Zamacois, Marc and Bellalta, Boris},
   year={2024},
   month=oct, pages={107919} }

@misc{Yeregui2024EdgeXR,
      title={Edge Rendering Architecture for multiuser XR Experiences and E2E Performance Assessment}, 
      author={Inhar Yeregui and Daniel Mejías and Guillermo Pacho and Roberto Viola and Jasone Astorga and Mario Montagud},
      year={2024},
      eprint={2406.07087},
      archivePrefix={arXiv},
      primaryClass={cs.NI},
      url={https://arxiv.org/abs/2406.07087}, 
}

@INPROCEEDINGS{Okafor2024MultiUserXR,
  author={Okafor, Okwudilichukwu and Esposito, Flavio and Pecorella, Tommaso},
  booktitle={2024 20th International Conference on Network and Service Management (CNSM)}, 
  title={EdgeVerse: Multi-User Virtual Reality via Edge Computing and eBPF}, 
  year={2024},
  volume={},
  number={},
  pages={1-4},
  doi={10.23919/CNSM62983.2024.10814299}}

@inproceedings{delmerico2018benchmark,
author = {Delmerico, Jeffrey and Scaramuzza, Davide},
title = {A Benchmark Comparison of Monocular Visual-Inertial Odometry Algorithms for Flying Robots},
year = {2018},
publisher = {IEEE Press},
url = {https://doi.org/10.1109/ICRA.2018.8460664},
doi = {10.1109/ICRA.2018.8460664},
pages = {2502–2509},
numpages = {8},
location = {Brisbane, Australia}
}

@misc{liu2024slamhive,
      title={Benchmarking SLAM Algorithms in the Cloud: The SLAM Hive Benchmarking Suite}, 
      author={Xinzhe Liu and Yuanyuan Yang and Bowen Xu and Delin Feng and Sören Schwertfeger},
      year={2024},
      eprint={2406.17586},
      archivePrefix={arXiv},
      primaryClass={cs.RO},
      url={https://arxiv.org/abs/2406.17586}, 
}

@inproceedings{sourya-vio,
author = {Saha, Sourya and Absur, Md. Nurul and Debroy, Saptarshi},
title = {Detection and Recovery of Adversarial Slow-Pose Drift in Offloaded Visual-Inertial Odometry},
year = {2025},
isbn = {9798400713538},
publisher = {Association for Computing Machinery},
address = {New York, NY, USA},
url = {https://doi.org/10.1145/3704413.3765307},
doi = {10.1145/3704413.3765307},
pages = {473–478},
numpages = {6},
location = {Rice University, Houston, TX, USA},
series = {MobiHoc '25}
}

@INPROCEEDINGS{effect,
  author={Zhang, Xiaojie and Pal, Amitangshu and Debroy, Saptarshi},
  booktitle={2021 IEEE/ACM 21st International Symposium on Cluster, Cloud and Internet Computing (CCGrid)}, 
  title={EFFECT: Energy-efficient Fog Computing Framework for Real-time Video Processing}, 
  year={2021},
  volume={},
  number={},
  pages={493-503},
  doi={10.1109/CCGrid51090.2021.00059}}

@INPROCEEDINGS{effect-dnn,
  author={Zhang, Xiaojie and Mounesan, Motahare and Debroy, Saptarshi},
  booktitle={2023 IEEE 24th International Symposium on a World of Wireless, Mobile and Multimedia Networks (WoWMoM)}, 
  title={EFFECT-DNN: Energy-efficient Edge Framework for Real-time DNN Inference}, 
  year={2023},
  volume={},
  number={},
  pages={10-20},
  doi={10.1109/WoWMoM57956.2023.00015}}

@INPROCEEDINGS{infer-edge,
  author={Mounesan, Motahare and Zhang, Xiaojie and Debroy, Saptarshi},
  booktitle={NOMS 2025-2025 IEEE Network Operations and Management Symposium}, 
  title={Infer-EDGE: Dynamic DNN Inference Optimization in Just-in-Time Edge-AI Implementations}, 
  year={2025},
  volume={},
  number={},
  pages={1-9},
  doi={10.1109/NOMS57970.2025.11073623}}

@inproceedings{xiaojiesec2023,
author = {Zhang, Xiaojie and Gan, Houchao and Pal, Amitangshu and Dey, Soumyabrata and Debroy, Saptarshi},
title = {On Balancing Latency and Quality of Edge-native Multi-view 3D Reconstruction},
year = {2024},
isbn = {9798400701238},
publisher = {Association for Computing Machinery},
address = {New York, NY, USA},
url = {https://doi.org/10.1145/3583740.3630267},
doi = {10.1145/3583740.3630267},
pages = {1–13},
numpages = {13},
location = {Wilmington, DE, USA},
series = {SEC '23}
}

@INPROCEEDINGS{motaharevec,
  author={Mounesan, Motahare and Lemus, Mauro and Yeddulapalli, Hemanth and Calyam, Prasad and Debroy, Saptarshi},
  booktitle={2024 IEEE 8th International Conference on Fog and Edge Computing (ICFEC)}, 
  title={Reinforcement Learning-driven Data-intensive Workflow Scheduling for Volunteer Edge-Cloud}, 
  year={2024},
  volume={},
  number={},
  pages={79-88},
  doi={10.1109/ICFEC61590.2024.00016}}

@inproceedings{xiaojie-short,
author = {Zhang, Xiaojie and Debroy, Saptarshi},
title = {Adaptive task offloading over wireless in mobile edge computing},
year = {2019},
isbn = {9781450367332},
publisher = {Association for Computing Machinery},
address = {New York, NY, USA},
url = {https://doi.org/10.1145/3318216.3363328},
doi = {10.1145/3318216.3363328},
booktitle = {Proceedings of the 4th ACM/IEEE Symposium on Edge Computing},
pages = {323–325},
numpages = {3},
location = {Arlington, Virginia},
series = {SEC '19}
}

@Article{yu-drl-xr-mobile-edge,
AUTHOR = {Yu, Xiaofan and Zhou, Siyuan and Wei, Baoxiang},
TITLE = {Dependent Task Offloading and Resource Allocation via Deep Reinforcement Learning for Extended Reality in Mobile Edge Networks},
JOURNAL = {Electronics},
VOLUME = {13},
YEAR = {2024},
NUMBER = {13},
ARTICLE-NUMBER = {2528},
URL = {https://www.mdpi.com/2079-9292/13/13/2528},
ISSN = {2079-9292},
DOI = {10.3390/electronics13132528}
}

@misc{asim2025generationimmersiveapplications5g,
      title={Towards Next Generation Immersive Applications in 5G Environments}, 
      author={Rohail Asim and Ankit Bhardwaj and Lakshmi Suramanian and Yasir Zaki},
      year={2025},
      eprint={2507.20050},
      archivePrefix={arXiv},
      primaryClass={cs.NI},
      url={https://arxiv.org/abs/2507.20050}, 
}

@article{BELLALTA2026104391,
title = {Understanding the Wi-Fi and VR streaming interplay: A comprehensible simulation and experimental study},
journal = {Journal of Network and Computer Applications},
volume = {245},
pages = {104391},
year = {2026},
issn = {1084-8045},
doi = {https://doi.org/10.1016/j.jnca.2025.104391},
url = {https://www.sciencedirect.com/science/article/pii/S1084804525002887},
author = {Boris Bellalta and Miguel Casasnovas and Ferran Maura and Alejandro Rodríguez and Juan S. Marquerie and Pablo L. García and Francesc Wilhelmi and Josep Blat}
}

@misc{git,
  author      = {GitHub},
  year        = {2026},
  title       = {Github repository},
  note        = {Accessed: Mar 14, 2026},
  howpublished= {\url{https://github.com/dissectlab/XR-CCGrid2026.git}}
}

@article{survey,
author = {Zhang, Xiaojie and Debroy, Saptarshi},
title = {Resource Management in Mobile Edge Computing: A Comprehensive Survey},
year = {2023},
issue_date = {December 2023},
publisher = {Association for Computing Machinery},
address = {New York, NY, USA},
volume = {55},
number = {13s},
issn = {0360-0300},
url = {https://doi.org/10.1145/3589639},
doi = {10.1145/3589639},
journal = {ACM Comput. Surv.},
month = jul,
articleno = {291},
numpages = {37}
}

\end{document}